\documentclass{article}

\usepackage[nonatbib,preprint]{neurips_2020}

\usepackage[utf8]{inputenc} \usepackage[T1]{fontenc}    \usepackage{hyperref}       \usepackage{url}            \usepackage{booktabs}       \usepackage{amsfonts}       \usepackage{nicefrac}       \usepackage{microtype}      

\usepackage{times}
\usepackage{latexsym}
\usepackage{graphicx}
\usepackage{amsmath}
\usepackage{amsfonts}
\usepackage{graphicx}
\usepackage{color}
\usepackage{url}
\usepackage{algorithm}
\usepackage[noend]{algpseudocode}
\usepackage{enumitem}
\usepackage[font={footnotesize}]{caption}

\usepackage{subcaption}
\usepackage{multirow}

\title{\modelname: Multi-Scale Representation Learning
over a Spherical Surface for Geospatial Predictions}

\author{Gengchen Mai* \\
  STKO Lab, Departemnt of Geography, UCSB
  \texttt{gengchen\_mai@ucsb.edu} \\
  \And
  Yao Xuan\thanks{The first two authors contribute equally to this work.} \\
  Department of Mathematics, UCSB \\
\texttt{yxuan@ucsb.edu } \\
  \And
  Wenyun Zuo \\
  Department of Biology, Stanford University \\
\texttt{wyzuo@stanford.edu} \\
  \And
  Krzysztof Janowicz \\
  STKO Lab, Departemnt of Geography, UCSB \\
\texttt{janowicz@ucsb.edu } \\
  \And
  Ni Lao \\
  SayMosaic Inc., Palo Alto, CA, USA, 94303 \\
\texttt{noon99@gmail.com}\\
}

\usepackage{amsthm}
\usepackage{amsmath}
\usepackage{mathabx}
\usepackage{xspace}
\usepackage{color}
\usepackage{pgfplots, pgfplotstable}
\usepackage[colorinlistoftodos]{todonotes}

\newcommand{\comment}[1]{}

\renewcommand{\vec}[1]{\boldsymbol{\mathbf{#1}}}

\def\mP{\mathcal{P}}
\def\Real{\mathbb{R}}

\def\lon{\lambda}
\def\lat{\phi}
\def\lons{\lon^s}
\def\lats{\lat^s}

\newcommand\latz[1]{\lat_{#1}}
\newcommand\lonz[1]{\lon_{#1}}

\def\lons{\lonz{s}}
\def\lats{\latz{s}}

\def\fs{f^s}
\def\th{\vec{x}}
\def\sd{\Delta D} \def\ca{\Delta\delta}

\newcommand{\nscale}{S}

\newcommand{\lr}{lr}

\newcommand{\numresnet}{h}
\newcommand{\numneuron}{k}

\newcommand{\maxscale}{r_{max}}
\newcommand{\minscale}{r_{min}}

\newcommand{\kernelsize}{\sigma}
\newcommand{\numkernel}{m}

\newcommand{\imgclsloss}{\mathcal{L}^{image}}
\newcommand{\lossweight}{\beta}
\newcommand{\sampleset}{\mathbb{X}}

\newcommand{\embdim}{d}

\newcommand{\peemb}{\mathbf{p}[\th]}

\newcommand{\negsamp}{\mathcal{N}}

\newcommand{\freq}{S}

\newcommand{\enc}{Enc}

\newcommand{\pemlp}{\mathbf{NN}}

\newcommand{\coordspasphere}{\mathbb{S}}
\newcommand{\params}{\theta}

\newcommand{\image}{\mathbf{I}}

\newcommand{\classemb}{\mathbf{T}}
\newcommand{\numclass}{c}
\newcommand{\classy}{y}

\newcommand{\act}{\sigma}

\newcommand{\pefunc}{f}

\newcommand{\tile}{tile}
\newcommand{\aodha}{wrap}

\newcommand{\rbf}{rbf}
\newcommand{\grid}{grid}
\newcommand{\theory}{theory}

\newcommand{\sphere}{sphereC}
\newcommand{\spheregrid}{sphereC+}
\newcommand{\spheremixscale}{sphereM}
\newcommand{\spheregridmixscale}{sphereM+}
\newcommand{\dft}{sphereDFS}

\newcommand{\modelname}{Sphere2Vec}

\newcommand{\spacevec}{Space2Vec}

\newcommand{\vspaceclustering}{\vspace*{-0.6cm}}
\newcommand{\vspacepred}{\vspace*{-0.5cm}}
\newcommand{\vspacepredd}{\vspace*{-0.7cm}}
\newcommand{\vspacecomp}{\vspace*{-0.4cm}}

\begin{document}

\maketitle

\begin{abstract}
Generating learning-friendly representations for points in a 2D space 
is a fundamental and long-standing problem in machine learning. 
Recently, multi-scale encoding schemes (such as 
\spacevec) were proposed to
directly encode any point in 2D space as a high-dimensional  vector, 
and has been successfully applied to various (geo)spatial prediction tasks.
However, a \textit{map projection distortion problem} rises when  applying location encoding models to large-scale real-world GPS coordinate datasets (e.g., species images taken all over the world) - \textit{all current location encoding models are designed for encoding points in a 2D (Euclidean) space but not on a spherical surface, e.g., earth surface}. 
To solve this problem, we propose a multi-scale location encoding model called $\modelname$ which directly encodes point coordinates on a spherical surface while avoiding the map projection distortion problem. 
We provide theoretical proof that the 
$\modelname$ encoding preserves the spherical surface distance between any two points. 
We also developed a unified view of distance-reserving encoding on spheres based on the Double Fourier Sphere (DFS).
We apply $\modelname$~ to the geo-aware image classification task.
Our analysis shows that  $\modelname$~ outperforms other 2D space location encoder models especially on the polar regions and data-sparse areas for image classification tasks because of its nature for spherical surface distance preservation.
\end{abstract}

\section{Introduction}
\label{sec:intro}

Location encoders \cite{chu2019geo,mac2019presence,mai2020multiscale,zhong2020reconstructing} refer to neural network architectures which encode a point in a 2D space (or 3D Euclidean space \cite{zhong2020reconstructing}) into a high dimensional embedding such that this kind of distributed representations are more learning-friendly for downstream machine learning models. Location encoders can be incorporated into the state-of-art models for many tasks to make them spatially explicit \cite{yan2019spatial,janowicz2020geoai}. In fact, location encoders have already shown promising performances on multiple tasks across different domains including geo-aware image classification \cite{chu2019geo,mac2019presence,mai2020multiscale}, POI classification \cite{mai2020multiscale}, trajectory prediction \cite{xu2018encoding}, geographic question answering \cite{mai2020se}, and 3D protein structure reconstruction \cite{zhong2020reconstructing}. 
Compared with well-established kernal-based approaches \cite{scholkopf2001kernel,xu2018encoding} such as Radius Based Function (RBF) which requires to memorize the training examples as the kernel centers for a robust prediction, inductive-learning-based location encoders \cite{chu2019geo,mac2019presence,mai2020multiscale,zhong2020reconstructing} have many advantages: 
1) they are more memory efficient since they do not need to memorize training samples; 
2) unlike RBF, the performance on unseen locations does not depend on the number and distribution of kernels.
Moreover, Gao et al. \cite{gao2018learning} have shown that grid like periodic representation can preserve absolute position information, relative distance and direction information in 2D Euclidean space. \cite{mai2020multiscale} further show that it benefit  the generalizability of down-steam models.

\begin{figure*}[t!]
	\centering \tiny
	\vspace*{-0.2cm}
	\begin{subfigure}[b]{0.99\textwidth}  
		\centering 
		\includegraphics[width=\textwidth]{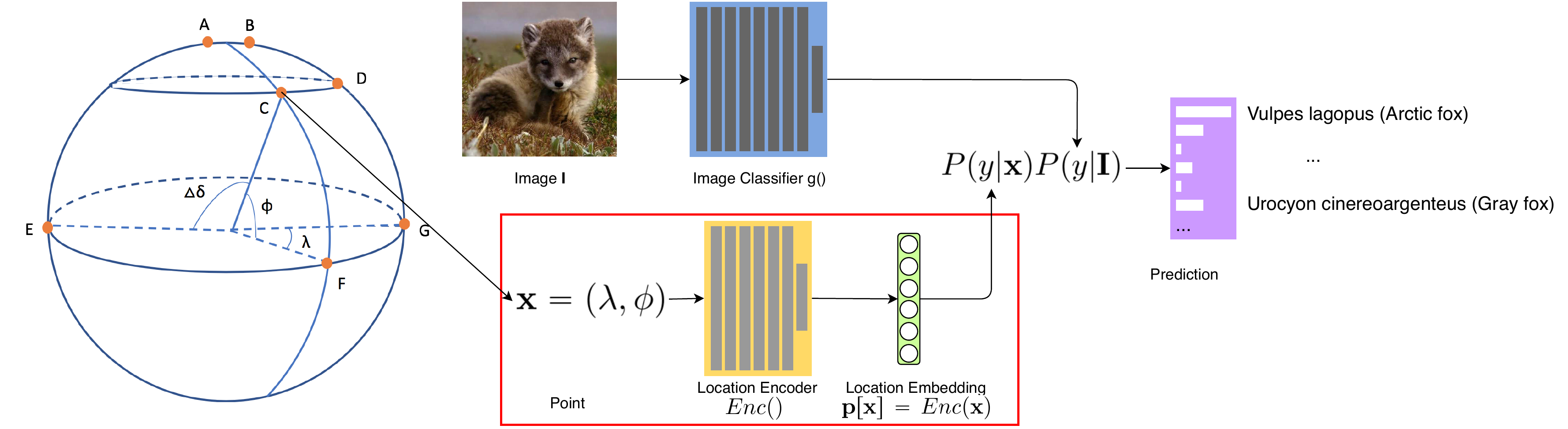} \vspace*{-0.2cm}
		\caption[]{{\small 
		$\modelname$ Illustration
		}}    
		\label{fig:pos_enc}
	\end{subfigure}
	\hfill
	\begin{subfigure}[b]{0.24\textwidth}  
		\centering 
		\includegraphics[width=\textwidth]{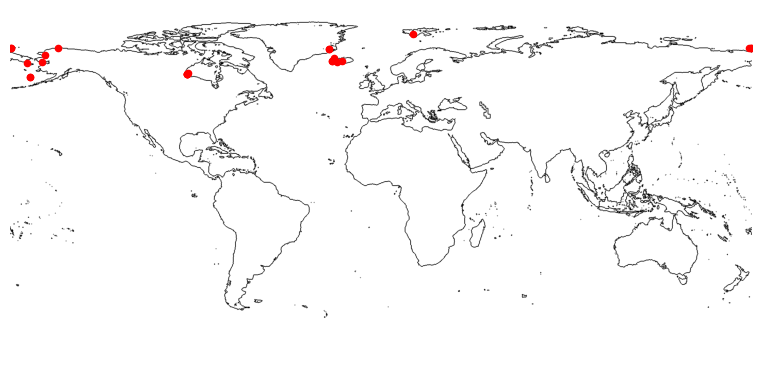}  \vspace*{-0.8cm} 
		\caption[]{{\small 
		Vulpes lagopus
		}}    
		\label{fig:4084_dist_intro}
	\end{subfigure}
	\hfill
	\begin{subfigure}[b]{0.24\textwidth}  
		\centering 
		\includegraphics[width=\textwidth]{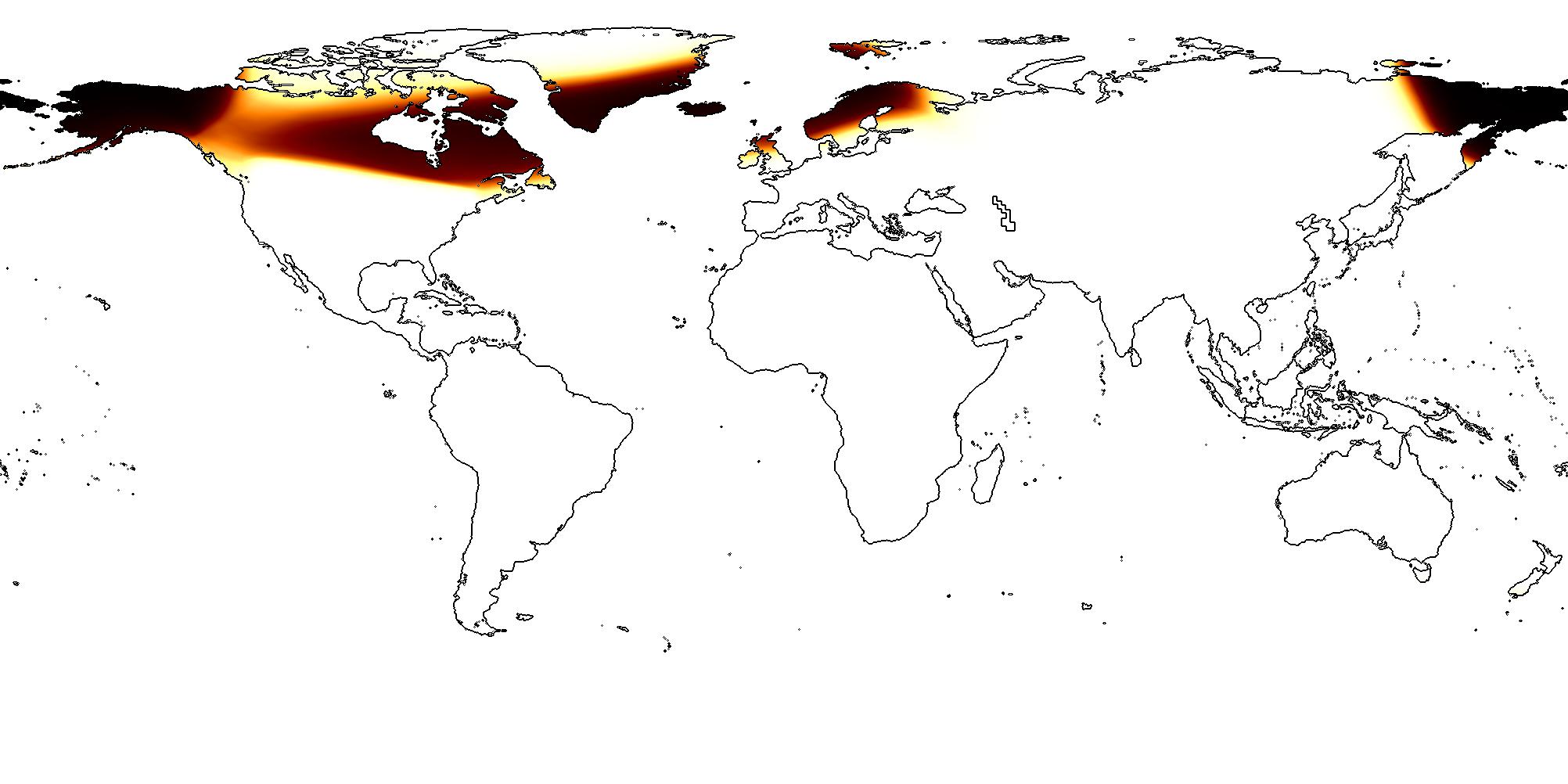}\vspacepred
		\caption[]{{\small 
		$\aodha*$
		}}    
		\label{fig:4084_aodha_intro}
	\end{subfigure}
	\hfill
	\begin{subfigure}[b]{0.24\textwidth}  
		\centering 
		\includegraphics[width=\textwidth]{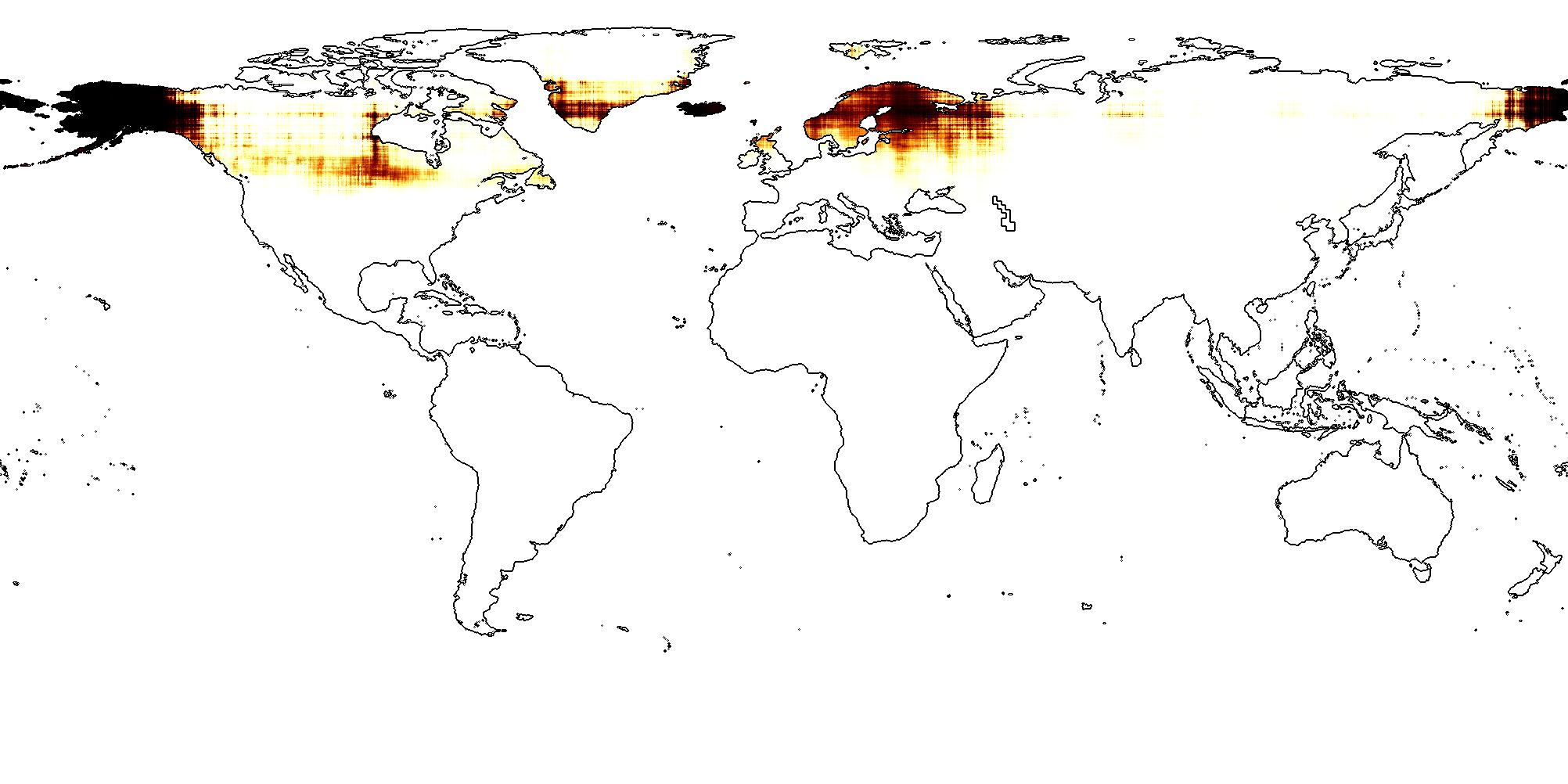}\vspacepred
		\caption[]{{\small 
		$\grid$
		}}    
		\label{fig:4084_grid_intro}
	\end{subfigure}
	\hfill
	\begin{subfigure}[b]{0.24\textwidth}  
		\centering 
		\includegraphics[width=\textwidth]{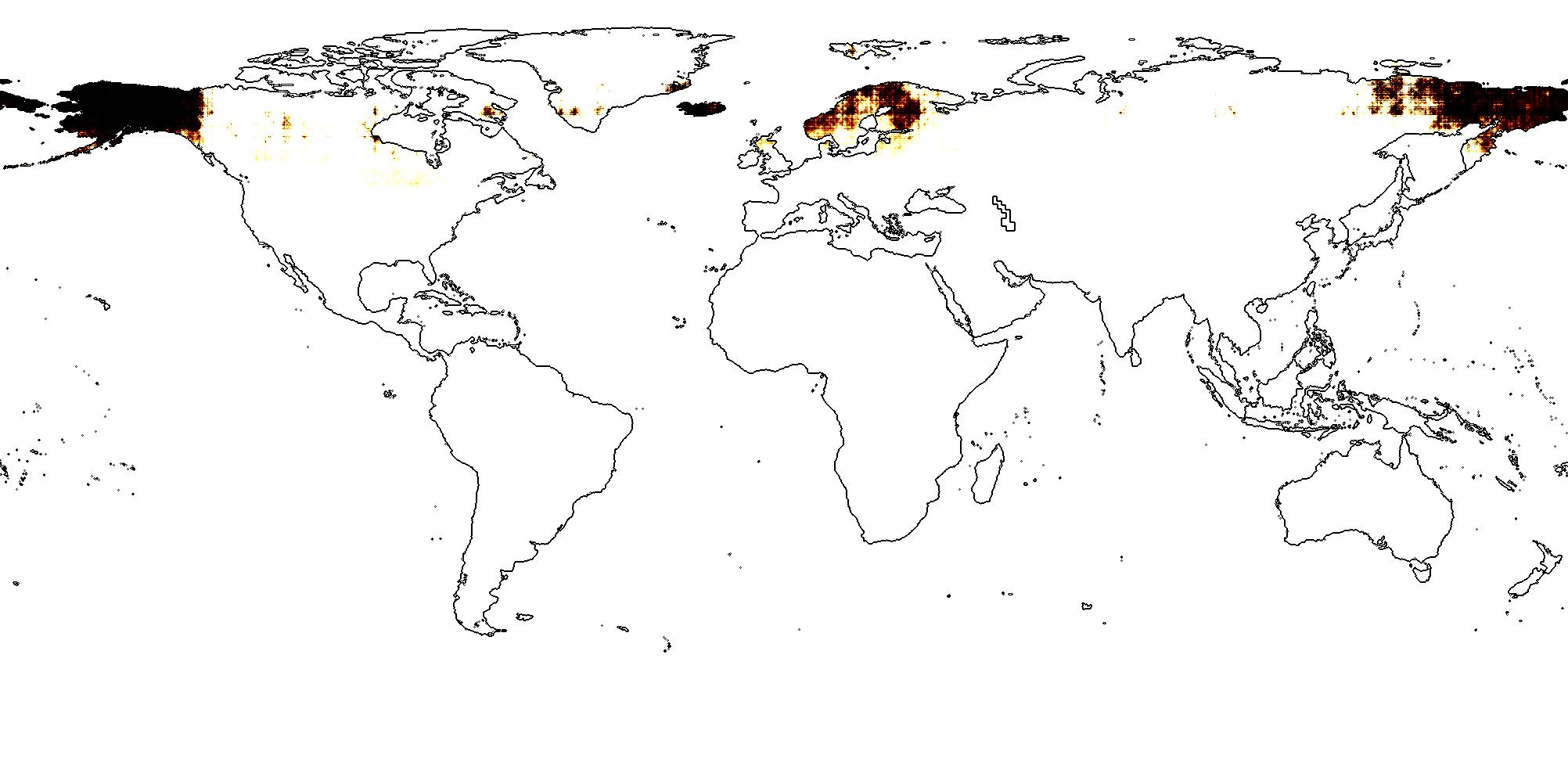}\vspacepred
		\caption[]{{\small 
		$\spheregrid$
		}}    
		\label{fig:4084_spheregrid_intro}
	\end{subfigure}
	\caption{(a)Applying $\modelname$ model for geo-aware image classification;
	(b): iNat2018 training image locations for Vulpes lagopus (Arctic fox).
	(c)-(e): The predicted  distributions of Vulpes lagopus from different models - $\aodha*$ \cite{mac2019presence}, $\grid$\cite{mai2020multiscale}, and $\spheregrid$ (our model). 
$\spheregrid$ produces a more compact and fine-grain distribution on the polar region while $\aodha*$ \cite{mac2019presence} turns to produce a over-generalized species distribution while $\grid$\cite{mai2020multiscale} is between two of them. For more examples, please see Figure \ref{fig:spesdist_exp}.
	} 
	\label{fig:spesdist_intro}
	\vspace*{-0.15cm}
\end{figure*}

Demand on  representation and prediction learning at a global scale grows dramatically due to emerging global scale issues, such as the transition path of the latest pandemic \cite{Chinazzi2020}, long lasting issue for malaria \cite{Caminade2014}, under threaten global biodiversity\cite{DiMarcoetal2019, Ceballosetal2020}, and numerous ecosystem and social system responses for climate change \cite{Hansen&Cramer2015}. 
However, when applying the state-of-the-art 2D space location encoders \cite{gao2018learning,chu2019geo,mac2019presence,mai2020multiscale,zhong2020reconstructing} to large-scale real-world GPS coordinate datasets such as species images taken all over the world, a \textbf{map projection distortion problem} \cite{williamson1973} emerges. These 2D location encoders  are designed for preserving distance in 2D (or 3D) Euclidean space, while GPS coordinates are in fact spherical coordinates, e.g., point $\th=(\lon,\lat)$ on a sphere with longitude $\lon \in  [-\pi , \pi)$ and latitude $\lat \in   [-{\pi}/{2} , -{\pi}/{2}]$ (See Figure \ref{fig:pos_enc}). 
Directly applying these 2D location encoders on spherical coordinates will yield a large distortion in the polar regions. It is important to point out that
\textit{map projection distortion is unavoidable when projecting spherical coordinates into 2D space.} (See Appendix \ref{sec:map_proj})
This emphasizes the importance of \textit{calculating on a round planet} \cite{chrisman2017calculating}.

Due to the limitations of these 2D location encoders, there is an urgent need for \textit{a location encoding method which preserves the spherical distance (e.g., great circle distance\footnote{\url{https://en.wikipedia.org/wiki/Great-circle_distance}}) between two points}.
In this work, we propose 
$\modelname$, which can directly encode point coordinates on a spherical surface while avoiding the map projection distortion. 
The mutli-scale encoding method utilizes 2D Discrete Fourier Transform\footnote{\url{http://fourier.eng.hmc.edu/e101/lectures/Image_Processing/node6.html}} basis ($O(S^2)$ terms) or a subset ($O(S)$ terms) of it while still being able to correctly measure the spherical distance.
We demonstrate the effectiveness of $\modelname$  on the geo-aware image classification (or so-call species fine-grain recognition) \cite{chu2019geo,mac2019presence,mai2020multiscale}. Following previous work we use location encoding to learn the geographic prior distribution of different species so that given an image and its associated location, we can combine the prediction of the location encoder and that from the state-of-the-art image classification models, e.g., inception V3 \cite{szegedy2016rethinking}, to improve the image classification accuracy. Figure \ref{fig:pos_enc} illsutrates the whole architecture. Given an image $\image$ in Figure \ref{fig:pos_enc}, its very difficult to decide whether it is an arctic fox or gray fox just based on the appearance information. However, if we know this image is taken from the Arctic area, then we have more confidence to say this is a baby arctic fox (Vulpes lagopus).
Figure \ref{fig:spesdist_intro} (c)-(e) shows the predicted distributions of \textit{Vulpes Lagopus} from three different models. We can see that $\modelname$~ has a clear advantage to capture fine-grained species distribution, especially on polar regions.
\textbf{In summary, the contributions of our work are:}
\begin{enumerate}
    \item We propose a multi-scale location encoder, $\modelname$, which, as far as we know,  is the first inductive embedding encoding scheme which aims at preserving spherical distance.
\item We provide a theoretical proof that 
$\modelname$~encodings can preserve spherical surface distance between points. 
    We also developed a unified view of distant reserving encoding methods on spheres based on Double Fourier Sphere (DFS) \cite{merilees1973,orszag1974}.
    \item We conduct extensive experiments on the geo-aware image classification task.
Results show that due to its distance preserving ability,  $\modelname$ outperforms the state-of-the-art 2D location encoder models. $\modelname$~ is able to produce  finer-grained and compact spatial distributions, and does significantly better on the polar regions and areas with sparse training samples.
\end{enumerate}

 \section{Problem Formulation}  \label{sec:prob}

{\em Distributed representation of point-features on the spherical surface} can be formulated as follows. 
Given a set of points $\mP=\{\th_i\}$ on the surface of a sphere $\coordspasphere^{2}$, e.g., species occurrences all over the world, where $\th_i=(\lon_i,\lat_i) \in \coordspasphere^{2}$ indicates a point with  longitude $\lon_i \in  [-\pi , \pi)$ and latitude $\lat_i \in   [-{\pi}/{2} , {\pi}/{2}]$.
Define a function $\enc_{\mP,\params}(\th): \coordspasphere^{2} \to \Real^\embdim$, 
which is parameterized by $\params$ and maps any coordinate $\th$ in a spherical surface $\coordspasphere^{2}$ to a vector representation of $\embdim$ dimension.

Let $ \enc(\th) = \pemlp(PE_{\freq}(\th))$ 
where $\pemlp()$ is a learnable multi-layer perceptron with $\numresnet$ hidden layers and $\numneuron$ neurons per layer. We want to find a function $PE_{\freq}(\th)$ which does a one-to-one mapping from each point $\th_i=(\lon_i,\lat_i) \in \coordspasphere^{2}$ to a multi-scale representation with $\freq$ as the total number of scales such that it satisfy the following requirement:
\begin{align}
    \langle PE_{\freq}(\th_1), PE_{\freq}(\th_2) \rangle = \pefunc(\sd), \forall \th_1, \th_2 \in \coordspasphere^{2},
    \label{equ:prob_stat}
\end{align}
where $\sd \in [0, \pi R]$ is the spherical surface distance between $\th_1, \th_2$,  $R$ is the radius of this sphere, and $\pefunc(x)$ is a strictly monotonically decreasing function for $x \in [0, \pi R]$. In other words, we expect to find a function $PE_{\freq}(\th)$ such that the resulting multi-scale representation of $\th$ preserves the spherical surface distance while it is more learning-friendly for the downstream neuron network model $\pemlp()$.

 \section{Related Work}
\label{sec:relatedwork}

\paragraph{Location Encoder}  \label{subsec:locenc_related}

There has been much research
on developing inductive learning-based location encoders. Most of them directly apply Multi-Layer Perceptron (MLP) to 2D coordinates to get a high dimensional location embedding for downstream tasks such as pedestrian trajectory prediction \cite{xu2018encoding} and geo-aware image classification
\cite{chu2019geo}.
Recently, Mac Adoha et al. \cite{mac2019presence} apply sinusoid functions to encode the latitude and longitude of each image
before feeding into MLPs.
All of the above approaches deploy encode locations at a single-scale.

Inspired by the position encoder in Transformer \cite{vaswani2017attention} and Neuroscience research on grid cells \cite{banino2018vector,cueva2018emergence} of mammals,  
Mai et al. \cite{mai2020multiscale} proposed to  apply multi-scale sinusoid functions to encode locations in 2D space before feeding into MLPs.
The multi-scale representations have advantages of capturing spatial feature distributions with different characteristics. 
Similarly, Zhong et al. \cite{zhong2020reconstructing} utilized a multi-scale location encoder for the position of proteins' atoms in 3D Euclidean space for protein structure reconstruction with great success. 
For a comprehensive survey of different location encoders, please refer to Mai et al. \cite{mai2021review}.

\paragraph{Machine Learning Models on Spheres}
\label{subsec:mapproj_related}
There has been much recent work on solving map projection distortion when designing machine learning models for large-scale real-world datasets.
For omnidirectional image classification task, 
both Cohen et al \cite{cohen2018spherical} and Coors et al. \cite{coors2018spherenet} design different spherical versions of the traditional CNN models in which the CNN filters explicitly considers map projection distortion. 
In terms of image geolocalization \cite{izbicki2019exploiting} and text geolocalization \cite{izbicki2019geolocating}, a loss function based on the mixture of von Mises-Fisher distributions (MvMF)-- a spherical analogue of the Gaussian mixture model (GMM)--  is used to replace the traditional cross-entropy loss for geolocalization models \cite{izbicki2019exploiting,izbicki2019geolocating}. 
All these works are closely related to geometric deep learning \cite{bronstein2017geometric}. They
show the importance to consider the spherical geometry, yet none of them considers representation learning in the embedding space.

\paragraph{Spatially-Explicit Machine Learning Models}  \label{subsec:spex_related}
There has been much work in
 \textit{improving the performance of 
machine learning models by
 using spatial features or spatial inductive bias} -- so called
spatially-explicit machine learning \cite{janowicz2020geoai,mai2021geographic}, or \textit{SpEx-ML}. 
 The spatial inductive bias 
in these models includes: 
 spatial dependency \cite{kejriwal2017neural,yan2019spatial}, 
 spatial heterogeneity \cite{berg2014birdsnap,chu2019geo,mac2019presence,mai2020multiscale}, 
 map projection \cite{cohen2018spherical,coors2018spherenet,izbicki2019exploiting,izbicki2019geolocating}, 
 scale effect \cite{weyand2016planet,mai2020multiscale}, and so on.

 \paragraph{Pseudospectral Methods on Spheres}  \label{subsec:pseudospectral}
Much study has been done for the numerical solutions on spheres, for example, in weather prediction 
\cite{orszag1972,orszag1974,merilees1973}.
The main idea is so called pseudospectral methods which leverage truncated discrete Fourier transformation on spheres to achieve computation efficiency while avoiding the error caused by projection distortion. 
The particular set of basis functions to be used depends on the particular problem.
However, they do not aim at learning good representations in machine learning models. In this study we try to make connections to these approaches and explore how their insights can be realized in a deep learning model. \section{Method} \label{sec:method}

In the following, we use $\enc(\th)$ to refer to $\enc_{\mP,\params}(\th)$. 
In Section \ref{subsec:encoder}, we will discuss our main contribution - the design of spherical distance-kept location encoder $\enc(\th)$, $\modelname$. 
We developed a unified view of distance-reserving encoding on spheres based on Double Fourier Sphere (DFS) \cite{merilees1973,orszag1974}.
Note that the resulting location embedding $\peemb  = \enc(\th)$ is a general-purpose embedding which can be utilized in different decoder architectures for various tasks. 
In Section \ref{subsec:img_cls_decoder}, we briefly show how to utilize the proposed $\enc(\th)$ in the geo-aware image classification task.

\subsection{\modelname}
\label{subsec:encoder}

Given any point $\th=(\lon,\lat) \in \coordspasphere^{2}$ with longitude $\lon \in  [-\pi , \pi)$ and latitude $\lat \in   [-\dfrac{\pi}{2} , \dfrac{\pi}{2}]$, the multi-scale location encoder proposed is in the form of $ \enc(\th) = \pemlp(PE_{\freq}(\th))$. 
$PE_{\freq}(\th)$ is a concatenation of multi-scale spherical spatial features of $S$ levels. In the following, we call $\enc(\th)$ {\it location encoder} and its component $PE_{\freq}(\th)$ {\it position encoder}.
Let $\minscale, \maxscale$ be the minimum and maximum  scaling factor, and $g = \frac{\maxscale}{\minscale}$.\footnote{In practice we fix $\maxscale=1$ meaning no scaling of $\lon,\lat$.}

\begin{figure*}[t!]
	\centering \tiny
	\vspace*{-0.2cm}
	\begin{subfigure}[b]{0.162\textwidth}  
		\centering 
		\includegraphics[width=\textwidth]{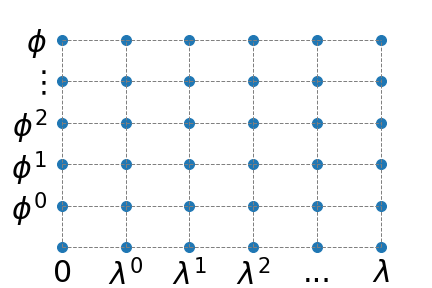}\vspace*{-0.2cm}
		\caption[]{{\small 
		$\dft$
		}}    
		\label{fig:dft}
	\end{subfigure}
	\begin{subfigure}[b]{0.162\textwidth}  
		\centering 
		\includegraphics[width=\textwidth]{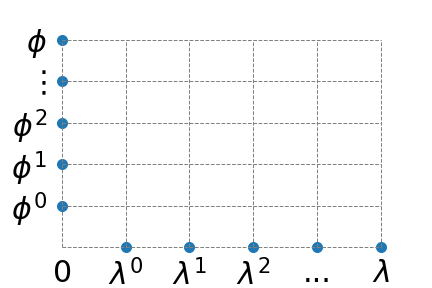}\vspace*{-0.2cm}
		\caption[]{{\small 
		$\grid$
		}}    
		\label{fig:grid}
	\end{subfigure}
	\begin{subfigure}[b]{0.162\textwidth}  
		\centering 
		\includegraphics[width=\textwidth]{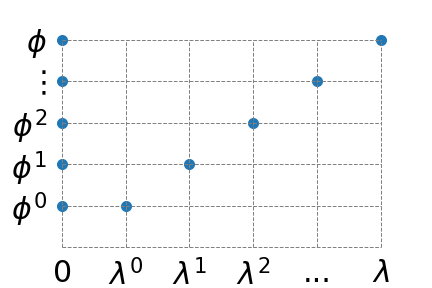}\vspace*{-0.2cm}
		\caption[]{{\small 
		$\sphere$
		}}    
		\label{fig:psphere}
	\end{subfigure}
	\begin{subfigure}[b]{0.162\textwidth}  
		\centering 
		\includegraphics[width=\textwidth]{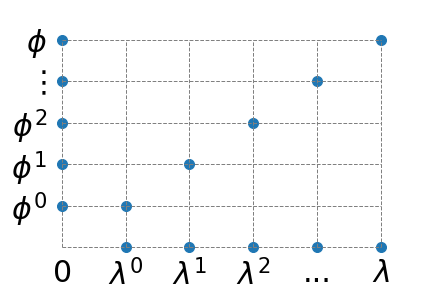}\vspace*{-0.2cm}
		\caption[]{{\small 
		$\spheregrid$
		}}    
		\label{fig:psheregrid}
	\end{subfigure}
	\begin{subfigure}[b]{0.162\textwidth}  
		\centering 
		\includegraphics[width=\textwidth]{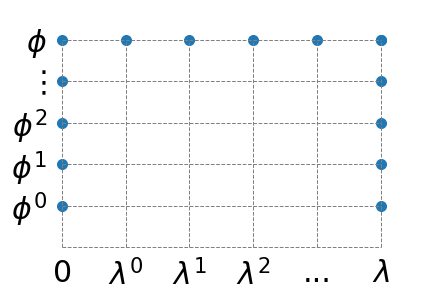}\vspace*{-0.2cm}
		\caption[]{{\small 
		$\spheremixscale$
		}}    
		\label{fig:psheremix}
		\end{subfigure}
		\begin{subfigure}[b]{0.162\textwidth}  
		\centering 
		\includegraphics[width=\textwidth]{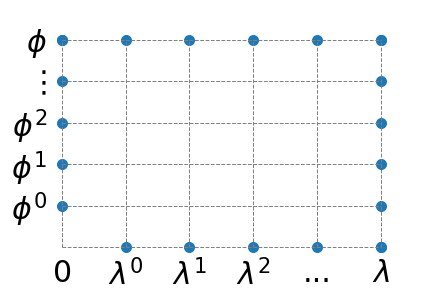}\vspace*{-0.2cm}
		\caption[]{{\small 
		$\spheregridmixscale$
		}}    
		\label{fig:psheremixgrid}
	\end{subfigure}

	\caption{Patterns of different encoders, blue points at $(\lon^m,\lat^n)$ mean interaction terms of trigonometric functions of $\lon^m$ and $\lat^n$ are included in the encoder, $\lon$ and $\lat$ axis correspond to single terms with no interactions.}
	\label{fig: allpatterns}
    \vspace*{-0.15cm}
\end{figure*}
 
\paragraph{\dft}
Double Fourier Sphere (DFS) \cite{merilees1973,orszag1974} is a simple yet successful pseudospectral method, which is computationally efficient and have been applied to analysis 
of large scale phenomenons such as weather \cite{sun2014} and blackholes \cite{bartnik2000}. Our first intuition is to use the base functions of DFS to help decompose $\th=(\lon,\lat)$ into a high dimensional vector:
\vspace{-0.1cm}
\begin{align} \begin{split}
   PE_{S,\dft}(\th)=&\bigcup_{n=0}^{\nscale-1}[\sin\latz{n}, \cos\latz{n}] \cup \bigcup_{m=0}^{\nscale-1} [\sin\lonz{m}, \cos\lonz{m}] \cup \\ 
    & \bigcup_{n=0}^{\nscale-1} \bigcup_{m=0}^{\nscale-1} [\cos\latz{n}\cos\lonz{m},  \cos\latz{n}\sin\lonz{m},
\sin\latz{n}\cos\lonz{m},\sin\latz{n}\sin\lonz{m}].
\label{equ:dfs}
\end{split}\end{align}
\vspace{-0.1cm}
where $\lons=\frac{\lon}{\fs}$, $\lats=\frac{\lat}{\fs}$, $\fs=\minscale \cdot g^{s/(\nscale-1)}$, $\cup$ means vector concatenation and $\bigcup_{s=0}^{\nscale-1}$ indicates vector concatenation through different scales.
It basically 
lets all the scales of $\lat$ terms interact with all the scales of $\lon$ terms in the encoder. This would introduce an encoder whose output has $O(S^2)$ dimensions and increase the memory burden in training and hurts generalization (Figure~\ref{fig: allpatterns}a). 
An encoder might achieve better result by only using a subset of these terms.
In comparison, the state of art encoder such as $\grid$ \cite{mai2020multiscale} and $\aodha$  \cite{mac2019presence} define: \footnote{As a multi-scale encoder, $\grid$ degenerates to $\aodha$ when the number of scales $S=1$.}
\begin{align}
    PE_{S,\grid}(\th)=\bigcup_{s=0}^{\nscale-1}[\sin\lats,\cos\lats,\sin\lons,\cos\lons].
\label{equ:grid}
\end{align}

We can see that $\grid$ employs a subset of terms from $\dft$ (Figure~\ref{fig: allpatterns}b). As we explained earlier, $\grid$ performs poorly at a global scale due to its inability to preserve spherical distances.
In the following we explore different subsets of DFS terms while achieving two goals: 1) efficient representation with $O(S)$ dimensions 2) preserving distance measures on spheres. 

\paragraph{\sphere}
Inspired by the fact that any point  $(x,y,z)$ in 3D Cartesian coordinate can be expressed by $sin$ and $cos$ basis of spherical coordinates ($\lon$, $\lat$ plus radius)
\footnote{\url{https://en.wikipedia.org/wiki/Spherical_coordinate_system}},
we define the basic form of \modelname, namely $\sphere$ encoder, for scale $s$ as
\begin{align}
   PE_{S,\sphere}(\th)=\bigcup_{s=0}^{\nscale-1}[\sin\lats,\cos\lats\cos\lons,\cos\lats\sin\lons].
    \label{equ:sphere}
\end{align}

To illustrate that $\sphere$ is good at capturing spherical distance, we take a close look at its basic case $S=1$ (define $s=0$ and $\fs=1$), where the multi-scale encoder degenerates to
\begin{align}
        PE_{1,\sphere}(\th)=[\sin(\lat),\cos(\lat)\cos(\lon),\cos(\lat)\sin(\lon)].
    \label{equ:sphere1}
\end{align}
These three terms are included in the multi-scale version ($S > 1$) and serve as the main terms at the largest scale and also the lowest frequency (when $s=S-1$). 
The high frequency terms are added to help the downstream neuron network to learn the point-feature more efficiently. 
Interestingly, $PE_{1,\sphere}$ captures the spherical distance in a very explicit way:
\newtheorem{theorem}{Theorem}
\begin{theorem}\label{thm1}
Let $\th_1$, $\th_2$ be two points on the same sphere with radius $R$, then 
\begin{align}
    \langle PE_{1,\sphere}(\th_1), PE_{1,\sphere}(\th_2) \rangle = \cos(\frac{\sd}{R}),
\end{align}
where $\sd$ is the great circle distance between $\th_1$ and $\th_2$. Under this metric,
\begin{align}
    \|PE_{1,\sphere}(\th_1)-PE_{1,\sphere}(\th_2)\|
= 2\sin (\frac{\Delta D}{2R}).
\end{align}
Moreover, $\|PE_{1,\sphere}(\th_1)-PE_{1,\sphere}(\th_2) \|\approx \frac{\sd}{R}$,when $\Delta D$ is small w.r.t. $R$.
\end{theorem}

\newtheorem{remark}{Remark}
See the proof in Section \ref{sec:proof1}.
Since the central angle $\ca = \frac{\Delta D}{R} \in [0, \pi]$ and $\cos(x)$ is strictly monotonically decrease for $x \in [0, \pi], $ Theorem $\ref{thm1}$ 
shows that $PE_{1,\sphere}(\th)$ directly satisfies our expectation in Equation \ref{equ:prob_stat} where $\pefunc(x) = \cos(\frac{x}{R})$. 
In comparison, the inner product in the output space of $\grid$ encoder is
\begin{align}
    \langle PE_{1,\grid}(\th_1), PE_{1,\grid}(\th_2) \rangle =\cos(\lat_1-\lat_2)+\cos(\lon_1-\lon_2),
\end{align}
which models the latitude difference and longitude difference of $\th_1$ and $\th_2$ separately rather than spherical distance.
This introduces problems in encoding.
For instance, consider data pairs $\th_1=(\lon_1,\lat)$ and $\th_2=(\lon_2,\lat)$, the distance between them in output space of $\grid$,  $\|PE_{1,\grid}(\th_1)-PE_{1,\grid}(\th_2)\|^2
= 2 - 2\cos(\lon_1-\lon_2)$ stays as a constant in terms of $\lat$. However,  when $\lat$ varies from $-\frac{\pi}{2}$ to $\frac{\pi}{2}$, the actual spherical distance changes in a wide range, e.g., the actual distance between the data pair at $\lat=-\frac{\pi}{2}$ (South Pole) is 0 while the distance between the data pair at $\lat=0$ (Equator), gets the maximum value.
This issue in measuring distances also has a negative impact on $\grid$'s ability to model distributions
in areas with sparse sample points because it is hard to learn the true spherical distance.
We observe that $\grid$ reaches peak performance at much smaller $\minscale$ than that of $\modelname$ encodings (See Appendix \ref{sec:param}). Moreover, $\sphere$ outperforms $\grid$ near polar regions where $\grid$ claims large distance though the spherical distance is small (A, B in Figure \ref{fig:pos_enc}).

\paragraph{\spheremixscale}
Considering the fact that many geographical features are more sensitive to either latitude (e.g., temperature, sunshine duration) or longitude (e.g., timezones, geopolitical borderlines),  we might want to focus on increasing the resolution of either  $\lat$ or $\lon$ while the other is hold relatively at large scale.  
Therefore, we introduce a  multi-scale position encoder  $\spheremixscale$, where interaction terms between $\lat$ and $\lon$ always have one of them fixed at top scale:
\begin{align}
\begin{split}
    PE_{S,\spheremixscale}(\th)=&\bigcup_{s=0}^{\nscale-1}[\sin\lats,\cos\lats\cos\lon,\cos\lat\cos\lons,\cos\lats\sin\lon,\cos\lat\sin\lons].\\
\end{split}\end{align}
This new encoder ensures that the $\lat$ term interact with all the scales of $\lon$ terms and $\lon$ term interact with all the scales of $\lat$ terms.  
Note that $PE_{1,\spheremixscale}$ is equivalent to $PE_{1,\sphere}$. Both $\sphere$ and $\spheremixscale$ are multi-scale versions of a spherical distance-kept encoder (See Equation \ref{equ:sphere1}) and keep that as the main term in their multi-scale representation.

\paragraph{\spheregrid~ and \spheregridmixscale}

From the above analysis of the two proposed position encoders and the state of art $\grid$ encoders, we know that $\grid$ pays more attention to the sum of $cos$ difference of latitudes and longitudes, while our proposed encoders pay more attention to the spherical distances. 
In order to capture both information, we consider merging $\grid$ with each proposed encoders to get more powerful models that encode geographical information from different angles. 
\begin{align}
  \label{eq:spherelasttwo}
 \begin{split}
        PE_{S,\spheregrid}(\th) &= PE_{S,\sphere}(\th) \cup PE_{S,\grid}(\th), \\
PE_{S,\spheregridmixscale}(\th)&= PE_{S,\spheremixscale}(\th) \cup PE_{S,\grid}(\th).
\end{split}
 \end{align}
We hypothesize that encoding these terms in the multi-scale representation would make the training of the encoder easier and the order of output dimension is still $O(S)$
(Figure \ref{fig: allpatterns}).

In location encoding, the uniqueness of encoding (no two points on sphere having the same position encoding) is very important, $PE_S(\th)$ in the five proposed methods are one-to-one mapping. 
\begin{theorem}
$\forall * \in \{\sphere,\spheregrid,\spheremixscale,\spheregridmixscale, \dft\}$, $PE_{S, *}(\th)$ is an injective function.
\label{the:injective}
\end{theorem}
See the proof in Section \ref{sec:proof3}.

\subsection{Applying \modelname~to Geo-Aware Image Classification}  \label{subsec:img_cls_decoder}

The \textit{geo-aware image classification task} \cite{chu2019geo,mac2019presence} can be formulated as:
Given an image $\image$ taken from location/point $\th$,  estimate which category $\classy$ it belongs to.
If we assume $\image$ and $\th$ are conditionally independent given $\classy$, then based on Bayes' theorem, we have $P(\classy|\image,\th) \; \propto \; P(\classy|\th)P(\classy|\image)$. $P(\classy|\image)$ can be given by any state-of-the-art image classification model such as Inception V3 \cite{szegedy2016rethinking}. In this work, we focus on estimating the geographic prior distribution of class $\classy$ over the spherical surface $P(\classy|\th) \; \propto \; \act(\enc(\th)\classemb_{:,\classy})$ where $\act(\dot)$ is a sigmoid activation function. 
$\classemb \in \Real^{\embdim \times \numclass}$ is a class embedding matrix where the $\classy_{th}$ column $\classemb_{:,\classy} \in \Real^{\embdim}$ indicates the class embedding for class $\classy$. Fig \ref{fig:pos_enc} illustrates the whole workflow. 
Please refer to Appendix \ref{subsec:img_cls_loss} and Mac Aodha et al. \cite{mac2019presence} for more details about this task, the loss function, and datasets.

 \section{Theoretical Proof of $\modelname$}  \label{sec:proof}

\subsection{Proof of Theorem 1} 
\label{sec:proof1}

\begin{proof}
	Since $PE_{1,\sphere}(\th_i)=[\sin(\lat_i),\cos(\lat_i)\cos(\lon_i),\cos(\lat_i)\sin(\lon_i)]$ for $i=1,2$, 
	the inner product
	\begin{align}
		\begin{split}
			&\langle PE_{1,\sphere}(\th_1), PE_{1,\sphere}(\th_2) \rangle\\
			&=\sin(\lat_1)\sin(\lat_2)+\cos(\lat_1)\cos(\lon_1)\cos(\lat_2)\cos(\lon_2)+\cos(\lat_1)\sin(\lon_1)\cos(\lat_2)\sin(\lon_2)\\
			&=\sin(\lat_1)\sin(\lat_2)+\cos(\lat_1)\cos(\lat_2)\cos(\lon_1-\lon_2)\\
			&=\cos(\ca)=\cos(\sd/R),
		\end{split}
	\end{align}
	where $\ca$ is the central angle between $\th_1$and $\th_2$, and the spherical law of cosines is applied to derive the second last equality. 
	So, 
	\begin{align}
		\begin{split}
			&\|PE_{1,\sphere}(\th_1)-PE_{1,\sphere}(\th_2)\|^2\\
			&=\langle PE_{1,\sphere}(\th_1)-PE_{1,\sphere}(\th_2),   PE_{1,\sphere}(\th_1)-PE_{1,\sphere}(\th_2) \rangle\\
			&=2-2\cos(\sd/R)\\
			&=4\sin^2(\sd/2R).
		\end{split}
	\end{align}
	
	So $\|PE_{1,\sphere}(\th_1)-PE_{1,\sphere}(\th_2)\|=2\sin(\sd/2R)$ since $\sd/2R\in [0,\frac{\pi}{2}]$.
	By Taylor expansion, $\|PE_{1,\sphere}(\th_1)-PE_{1,\sphere}(\th_2)\|\approx \sd/R$ when $\sd$ is small w.r.t. $R$.
\end{proof}

\subsection{Proof of Theorem \ref{the:injective}}
\label{sec:proof3}

$\forall * \in \{\sphere,\spheregrid,\spheremixscale,\spheregridmixscale\}$, $PE_{S,*}(\th_1)=PE_{S,*}(\th_2)$ implies 
\begin{align}
	\sin(\lat_1)=\sin(\lat_2),
	\label{z1=z2}
\end{align}
\begin{align}
	\cos(\lat_1)\sin(\lon_1)=\cos(\lat_2)\sin(\lon_2),
	\label{y1=y2}
\end{align}
\begin{align}
	\cos(\lat_1)\cos(\lon_1)=\cos(\lat_2)\cos(\lon_2),
	\label{x1=x2}
\end{align}
from $s=0$ terms.
Equation \ref{z1=z2} implies $\lat_1=\lat_2$. If $\lat_1=\lat_2=\pi/2$, then both points are at North Pole, $\lon_1=\lon_2$ equal to whatever longitude defined at North Pole. If $\lat_1=\lat_2=-\pi/2$, it is similar case at South Pole. When $\lat_1=\lat_2\in (-\frac{\pi}{2},\frac{\pi}{2})$, $\cos (\lat_1)=\cos(\lat_2)\neq 0$. Then from Equation  \ref{y1=y2} and \ref{x1=x2}, 
\begin{align}
	\sin{\lon_1}=\sin(\lon_2),
	\cos(\lon_1)=\cos(\lon_2),
\end{align}
which shows that $\lon_1=\lon_2$. In summary, $\th_1=\th_2$, so $PE_{S,*}$ is injective.\\
If $*=\dft$, $PE_{S,*}(\th_1)=PE_{S,*}(\th_2)$ implies  
\begin{align}
	\sin(\lat_1)=\sin(\lat_2),
	\cos(\lat_1)=\cos(\lat_2),
	\sin(\lon_1)=\sin(\lon_2),
	\cos(\lon_1)=\cos(\lon_2),
\end{align} 
which proves $\th_1=\th_2$ and $PE_{S,*}$ is injective directly.  \section{Experiment}
\label{exp}

\subsection{Geo-Aware Image Classification}  \label{sucsec:img_cls_exp}

\begin{table}[ht!]
    \caption{Geo-aware image classification results.
The results of the first six baseline models are  from Mac Aodha et al. 
\cite{mac2019presence}. 
$\aodha*$ is the best results we obtained when rerunning the code provided by Mac Aodha et al. \cite{mac2019presence} while $\aodha$ indicates the original results reported by Mac Aodha et al. \cite{mac2019presence}. 
Since the test sets for iNat2017 and iNat2018 are not open-sourced, we report results on validation sets.
The best performance of the baseline models and $\modelname$ are highlighted as bold.
See Appendix \ref{sec:param} for  $\modelname$ hyperparameters.
}
	\label{tab:imgcls_eval}
	\centering
{	\small 
\begin{tabular}{l|c|c|c|c|c|c}
\toprule
& BirdSnap & BirdSnap$\dagger$ & NABirds$\dagger$ & iNat2017 & iNat2018 & Avg\\ \hline
P(y|x) prior model 
& Test & Test & Test & Val & Val & - \\ \hline
No Prior (i.e. image model) & 70.07 & 70.07 & 76.08 & 63.27 & 60.20 & 67.94 \\ \hline
Nearest Neighbor (num) & 70.82 & 77.76 & 79.99 & 65.35 & 68.72 & 72.53 \\
Nearest Neighbor (spatial) & 71.57 & 77.98 & 80.79 & 65.55 & 67.51 & 72.68 \\
$\tile$
\cite{tang2015improving} (location only) & 71.13 & 77.19 & 79.58 & 65.49 & 67.02 & 72.08 \\
Adaptive Kernel \cite{berg2014birdsnap} & 71.57 & 78.65 & 81.11 & 64.86 & 65.23 & 72.28 \\
$\aodha$ \cite{mac2019presence} (location only)& 71.66 & 78.65 & 81.15 & \textbf{69.34} & 72.41 & 74.64 \\
$\aodha*$ \cite{mac2019presence} (location only) & 71.71 & 78.78 & 81.23 & 69.17 & 72.51 & 74.68  \\  $\rbf$\cite{mai2020multiscale} 
& 71.04 & 78.56 & 81.13 & 66.87 & 70.31 & 73.58 \\
$\grid$ \cite{mai2020multiscale} 
& 71.62 & \textbf{79.44} & 81.28 & 69.10 & \textbf{72.80} & \textbf{74.85} \\
$\theory$ \cite{mai2020multiscale} 
& \textbf{71.97} & 79.35 & \textbf{81.59} & 68.45 & 72.79 & 74.83 \\ \hline
$\sphere$
& 71.97 & 79.75 & 81.66 & 69.62 & 73.29 & 75.26 \\
$\spheregrid$ & \textbf{72.41} & 80.06 & 81.66 & \textbf{69.70} & \textbf{73.31} & \textbf{75.43} \\
$\spheremixscale$ & 72.02 & 79.75 & 81.65 & 69.69 & 73.25 & 75.27 \\
$\spheregridmixscale$ & 71.88 & \textbf{80.11} & \textbf{81.68} & 69.67 & 73.27 & 75.32 \\
$\dft$ & 71.75 & 79.18 & 81.31 & 69.65 & 73.24 & 75.03 \\
\bottomrule
\end{tabular} }
\end{table}
 
We conduct experiments on five  large-scale real-world datasets. Please refer to Mac Aodha et al. \cite{mac2019presence} for the dataset description. Table \ref{tab:imgcls_eval} compares the evaluation result among our 5 $\modelname$~ models with multiple baseline models. We can see that our first 4 $\modelname$~ models outperform all baselines on all five datasets except that $\spheregridmixscale$ are slightly worse that $\theory$ on BirdSnap. This clearly show the advantages of $\modelname$~ to handle large-scale geographic datasets. 

\begin{figure*}[ht!]
	\centering \small \begin{subfigure}[b]{0.41\textwidth}  
		\centering 
		\includegraphics[trim=0cm 1.5cm 0cm 0cm, clip=true, width=\textwidth, height=1.2in]{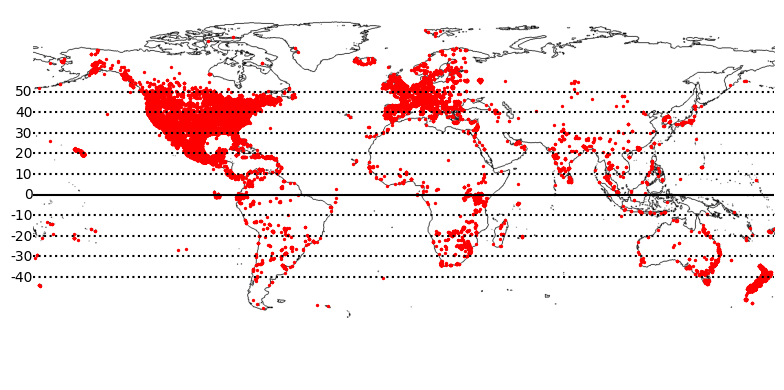}
		\vspacecomp
		\caption[]{{\small 
		Validation Locations
		}}    
		\label{fig:inat17_locs}
	\end{subfigure}
	\hfill
	\begin{subfigure}[b]{0.28\textwidth}  
		\centering 
		\includegraphics[width=\textwidth]{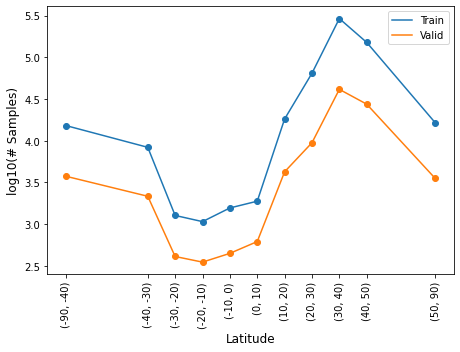}
		\vspacecomp
		\caption[]{{\small 
Samples per $\lat$ band
		}}    
		\label{fig:inat17_num_sample}
	\end{subfigure}
	\hfill
	\begin{subfigure}[b]{0.28\textwidth}  
		\centering 
		\includegraphics[width=\textwidth]{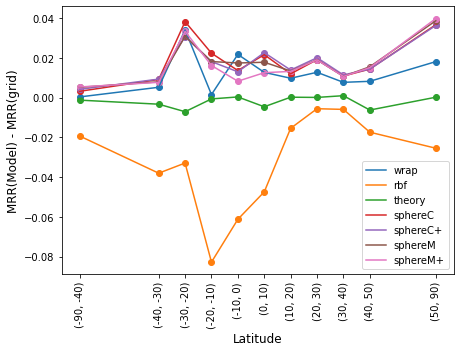}
		\vspacecomp
		\caption[]{{\small 
		$\Delta MRR$ per $\lat$ band
		}}    
		\label{fig:inat17_mrr}
	\end{subfigure}
	
	\hfill
	\begin{subfigure}[b]{0.41\textwidth}  
		\centering \includegraphics[trim=0cm 1cm 0cm 0.5cm, clip=true, width=\textwidth, height=1.1in]{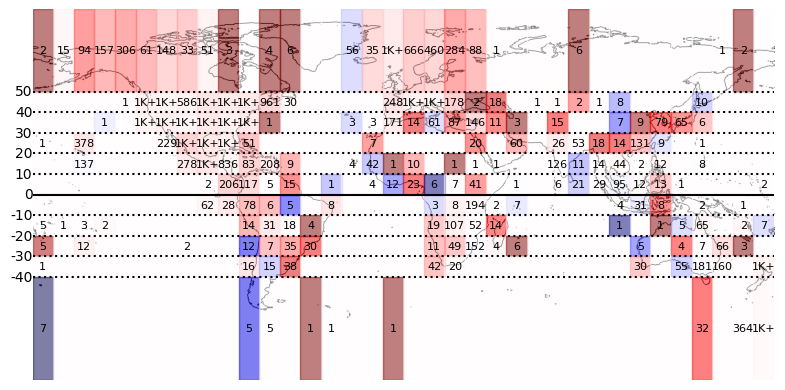}
		\vspacecomp
		\caption[]{{\small 
		$\Delta MRR$ per cell
		}}    
		\label{fig:inat17_cell_mrr}
	\end{subfigure}
	\hfill
	\begin{subfigure}[b]{0.28\textwidth}  
		\centering 
		\includegraphics[width=\textwidth]{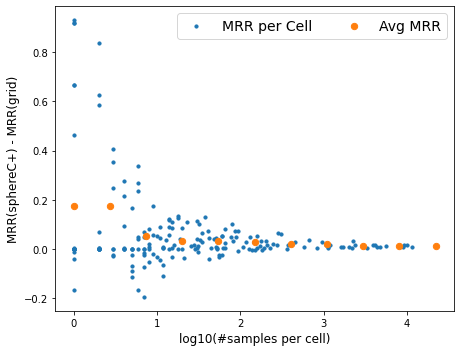}
		\vspacecomp
		\caption[]{{\small 
$\Delta MRR$ per cell
		}}    
		\label{fig:inat17_cell_scatter}
	\end{subfigure}
	\hfill
	\begin{subfigure}[b]{0.28\textwidth}  
		\centering 
		\includegraphics[width=\textwidth]{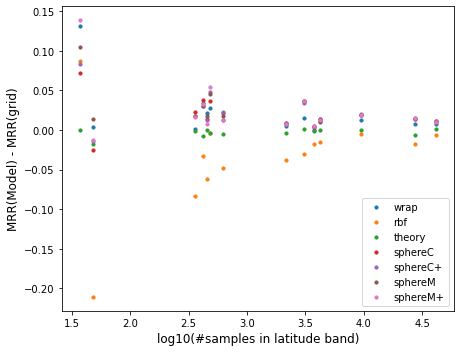}
		\vspacecomp
		\caption[]{{\small 
$\Delta MRR$ per $\lat$ band
		}}    
		\label{fig:inat17_band_scatter}
	\end{subfigure}
	\caption{iNat2017 dataset  and model performance comparison:
(a) Sample locations for validation set where the dashed and solid lines indicates latitudes; 
	(b) The number of training and validation samples in different latitude intervals.
(c) $MRR$ difference between a model and baseline $\grid$ on the validation dataset.
	(d) $\Delta MRR = MRR(\spheregrid) - MRR(grid)$ for each latitude-longitude cell. Red and blue color indicates positive and negative $\Delta MRR$ while darker color means high absolute value. The number on each cell indicates the number of validation data points  while "1K+" means there are more than 1K points in a cell.
	(e) The number of validation samples v.s. $\Delta MRR = MRR(\spheregrid) - MRR(grid)$ per latitude-longitude cell. The orange dots represent moving averages.
	(b) The number of validation samples v.s. $\Delta MRR$ per latitude band.
} 
	\label{fig:inat17_analysis}
	\vspace*{-0.15cm}
\end{figure*}

In order to further understand the reason for the superiority of $\modelname$~, we use iNat2017 and iNat2018 to conduct multiple analysis since they are the most up-to-date and have better coverage among these five datasets. Figure \ref{fig:inat17_locs} shows the image locations in iNat2017 validation dataset. We split this dataset into different latitude bands and compare the $\Delta MRR$ between each model to $\grid$. Figure \ref{fig:inat17_num_sample} and \ref{fig:inat17_mrr} show the number of samples and $\Delta MRR$ in each band while Figure \ref{fig:inat17_band_scatter} shows that the contrast between these two variables for different models. We can clearly see that 4 $\modelname$~ have larger $\Delta MRR$ are the North Pole (See Figure \ref{fig:inat17_mrr}). Moveover, $\modelname$~ has advantages on bands with less data samples, e.g. $\lat \in [-30^{\circ}, -20^{\circ})$. To have more concrete understanding, we compare $\spheregrid$ and $\grid$ in a higher spatial resolution - in latitude-longitude cells - which can be seen in Figure \ref{fig:inat17_cell_mrr} and \ref{fig:inat17_cell_scatter}. We can see that $\spheregrid$ have more advantages over $\grid$ in North Pole and data sparse cells.

\begin{figure*}[t!]
	\centering \tiny
	\vspace*{-0.2cm}
\begin{subfigure}[b]{0.24\textwidth}  
		\centering 
		\includegraphics[width=\textwidth]{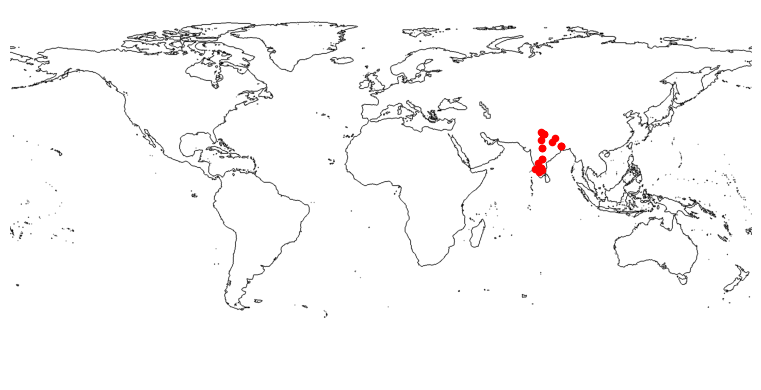}
		\vspacepredd
		\caption[]{{\scriptsize 
		Motacilla maderaspatensis
		}}    
		\label{fig:3475_dist_exp}
	\end{subfigure}
	\hfill
	\begin{subfigure}[b]{0.24\textwidth}  
		\centering 
		\includegraphics[width=\textwidth]{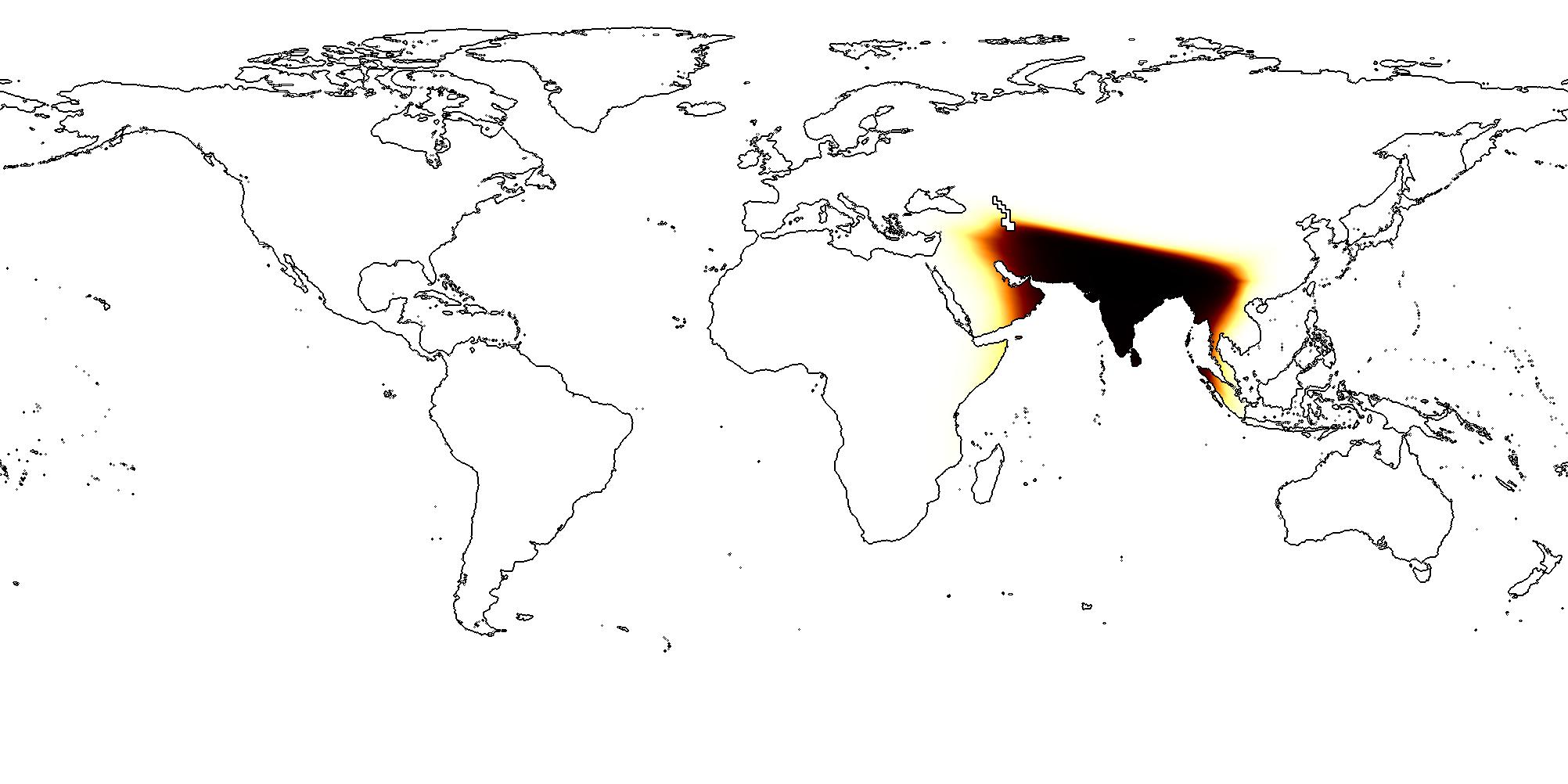}
		\vspacepredd
		\caption[]{{\small 
		$\aodha*$
		}}    
		\label{fig:3475_aodha_exp}
	\end{subfigure}
	\hfill
	\begin{subfigure}[b]{0.24\textwidth}  
		\centering 
		\includegraphics[width=\textwidth]{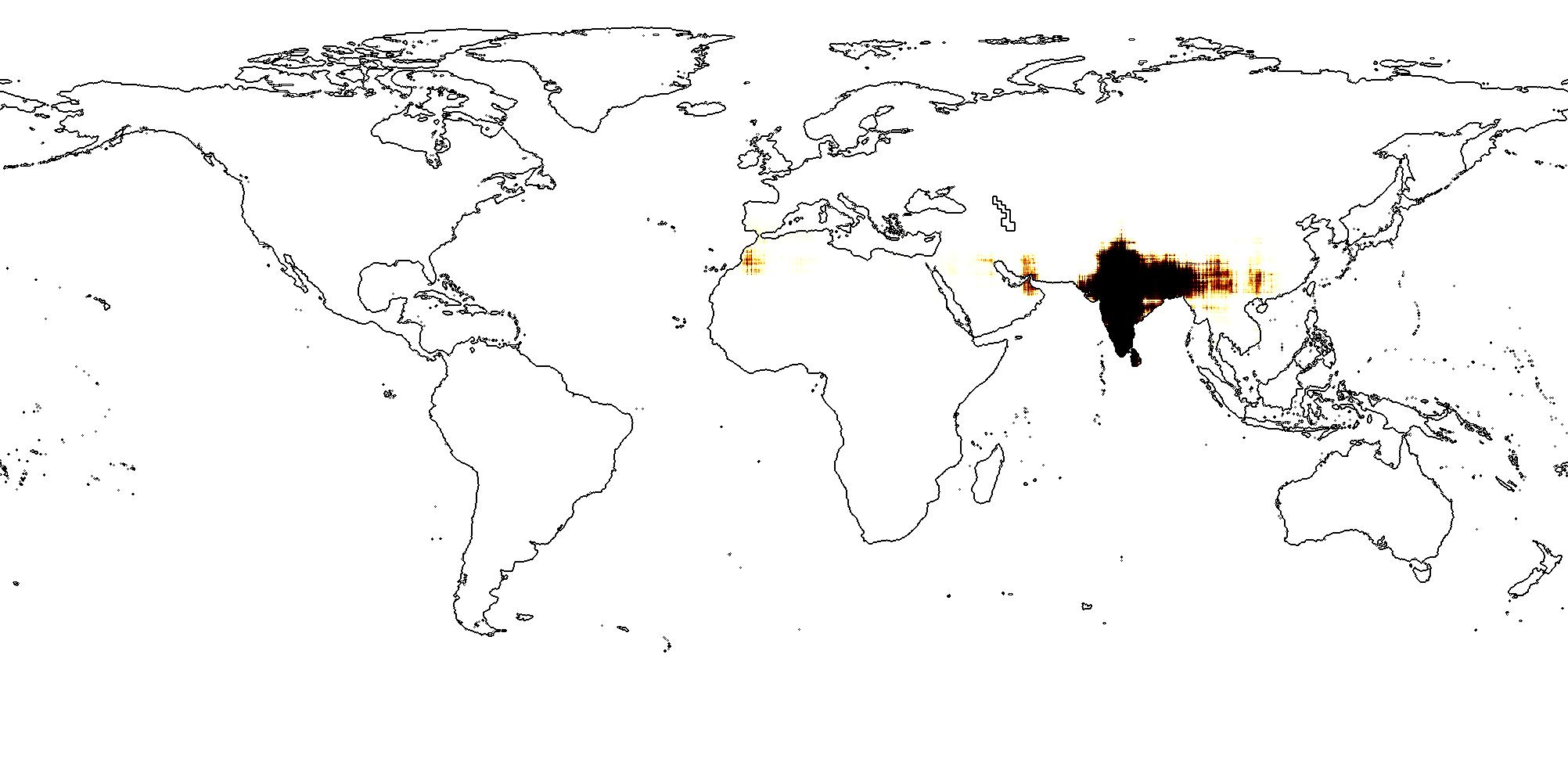}
		\vspacepredd
		\caption[]{{\small 
		$\grid$
		}}    
		\label{fig:3475_grid_exp}
	\end{subfigure}
	\hfill
	\begin{subfigure}[b]{0.24\textwidth}  
		\centering 
		\includegraphics[width=\textwidth]{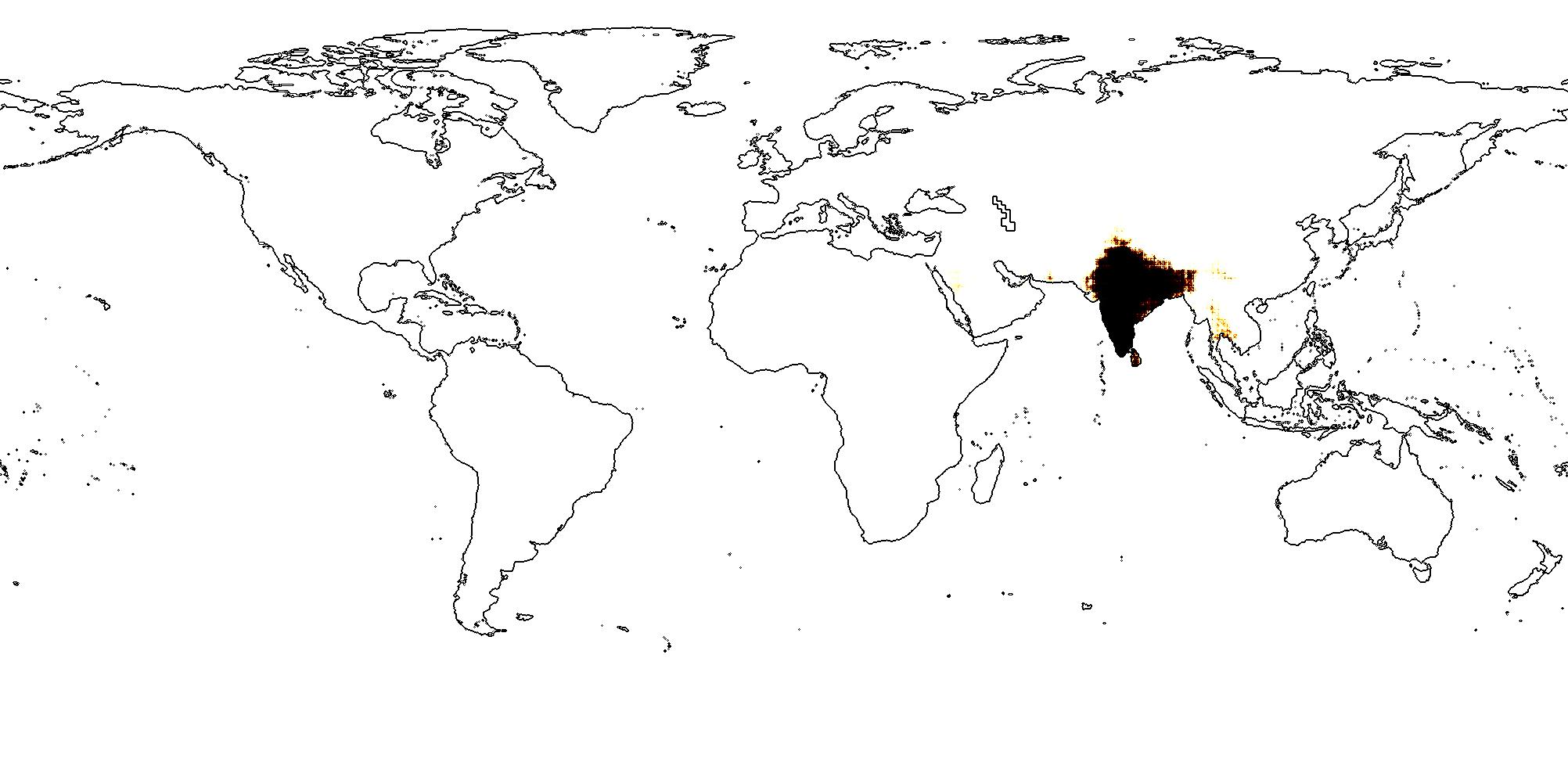}
		\vspacepredd
		\caption[]{{\small 
		$\spheregrid$
		}}    
		\label{fig:3475_spheregrid_exp}
	\end{subfigure}
\hfill
	\begin{subfigure}[b]{0.24\textwidth}  
		\centering 
		\includegraphics[width=\textwidth]{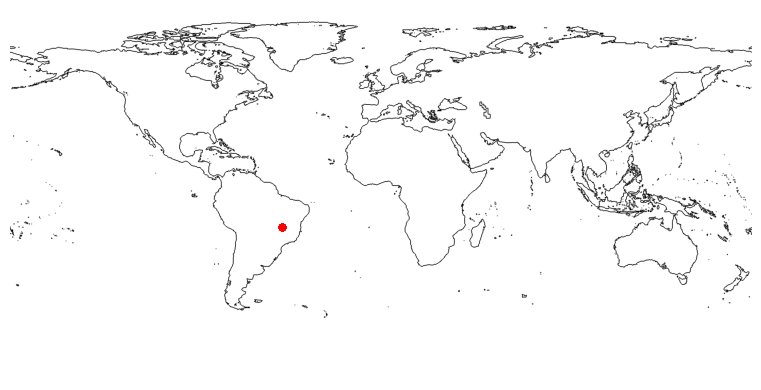}
		\vspacepredd
		\caption[]{{\small 
		Siderone galanthis
		}}    
		\label{fig:1555_dist_exp}
	\end{subfigure}
	\hfill
	\begin{subfigure}[b]{0.24\textwidth}  
		\centering 
		\includegraphics[width=\textwidth]{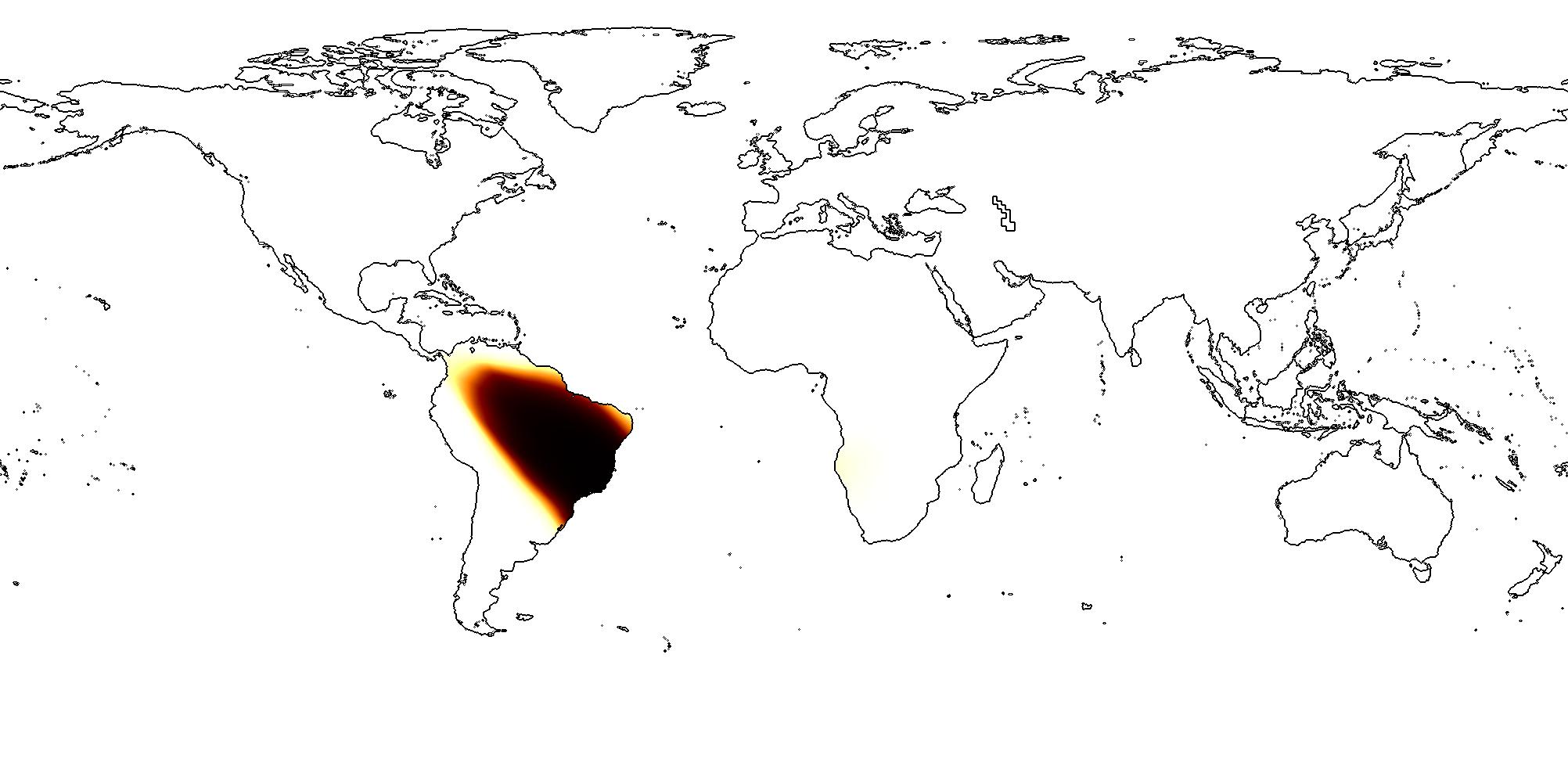}
		\vspacepredd
		\caption[]{{\small 
		$\aodha*$
		}}    
		\label{fig:1555_aodha_exp}
	\end{subfigure}
	\hfill
	\begin{subfigure}[b]{0.24\textwidth}  
		\centering 
		\includegraphics[width=\textwidth]{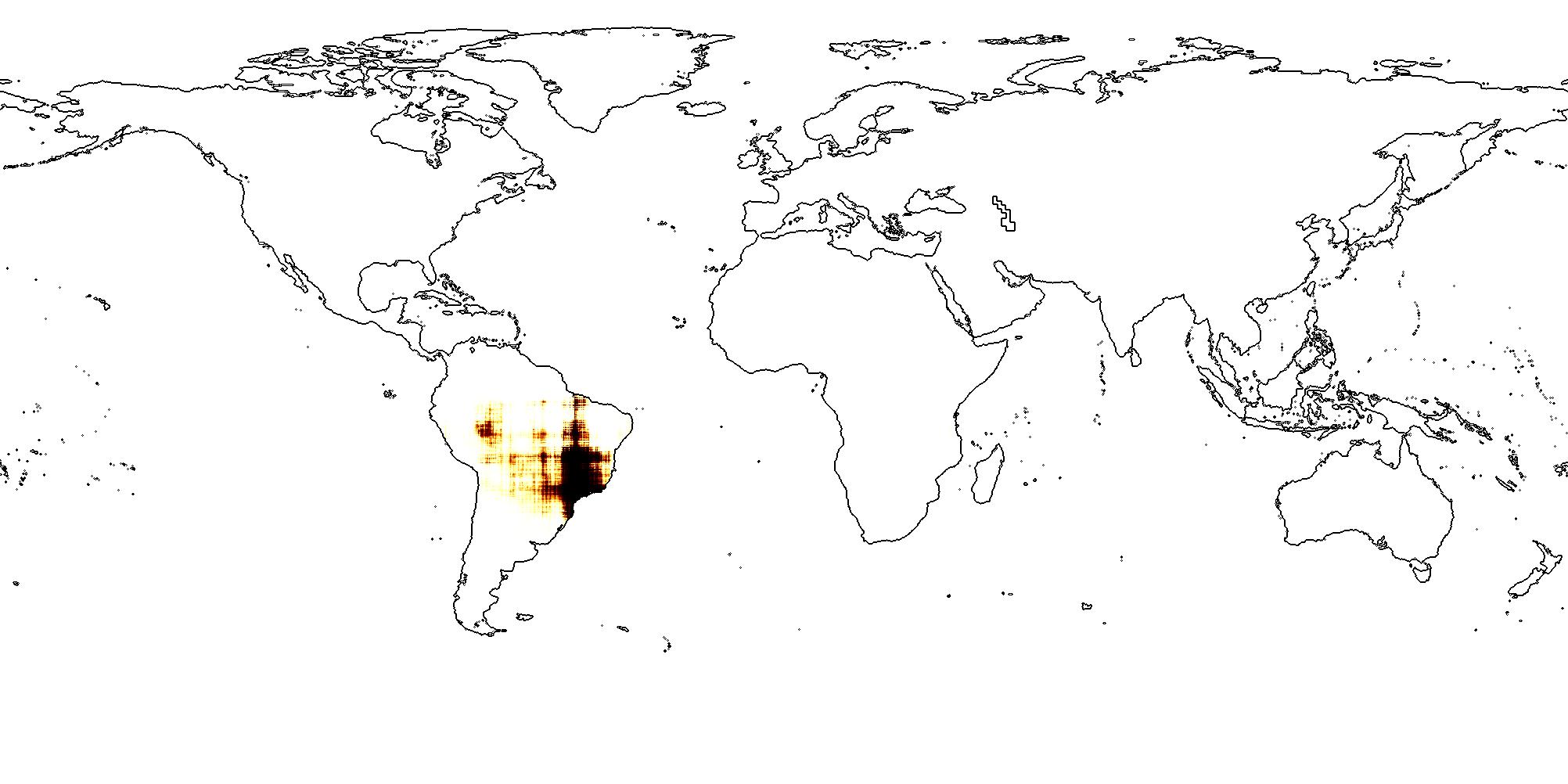}
		\vspacepredd
		\caption[]{{\small 
		$\grid$
		}}    
		\label{fig:1555_grid_exp}
	\end{subfigure}
	\hfill
	\begin{subfigure}[b]{0.24\textwidth}  
		\centering 
		\includegraphics[width=\textwidth]{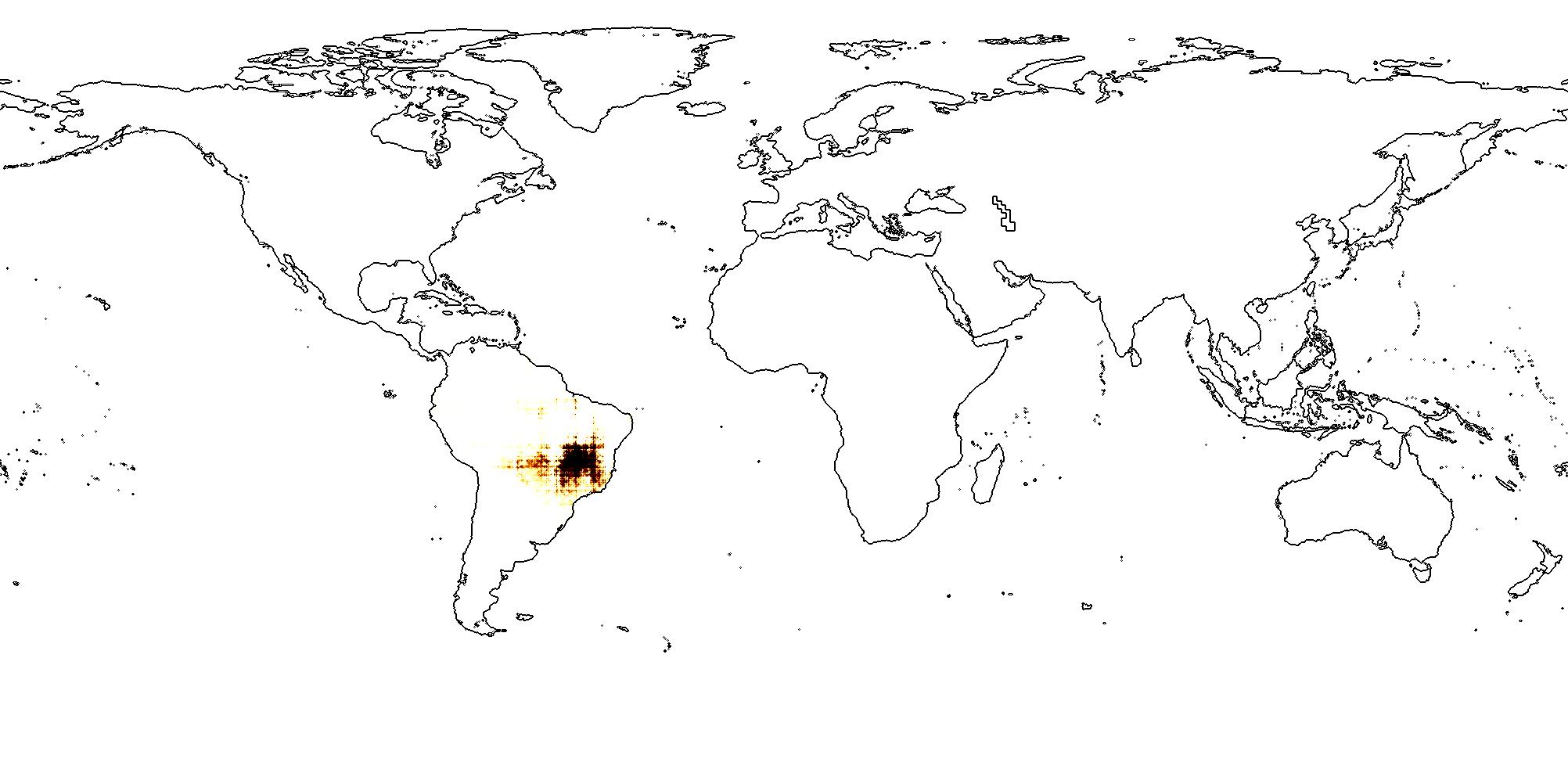}
		\vspacepredd
		\caption[]{{\small 
		$\spheregrid$
		}}    
		\label{fig:1555_spheregrid_exp}
	\end{subfigure}
	\hfill
	\begin{subfigure}[b]{0.24\textwidth}  
		\centering 
		\includegraphics[width=\textwidth]{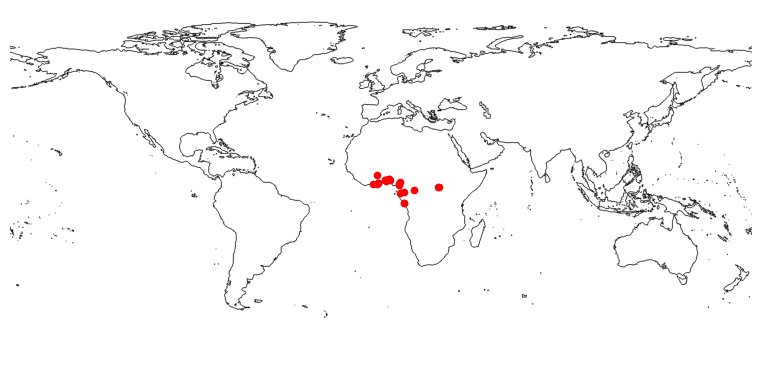}
		\vspacepredd
		\caption[]{{\small 
		Lophoceros fasciatus
		}}    
		\label{fig:2891_dist_exp}
	\end{subfigure}
	\hfill
	\begin{subfigure}[b]{0.24\textwidth}  
		\centering 
		\includegraphics[width=\textwidth]{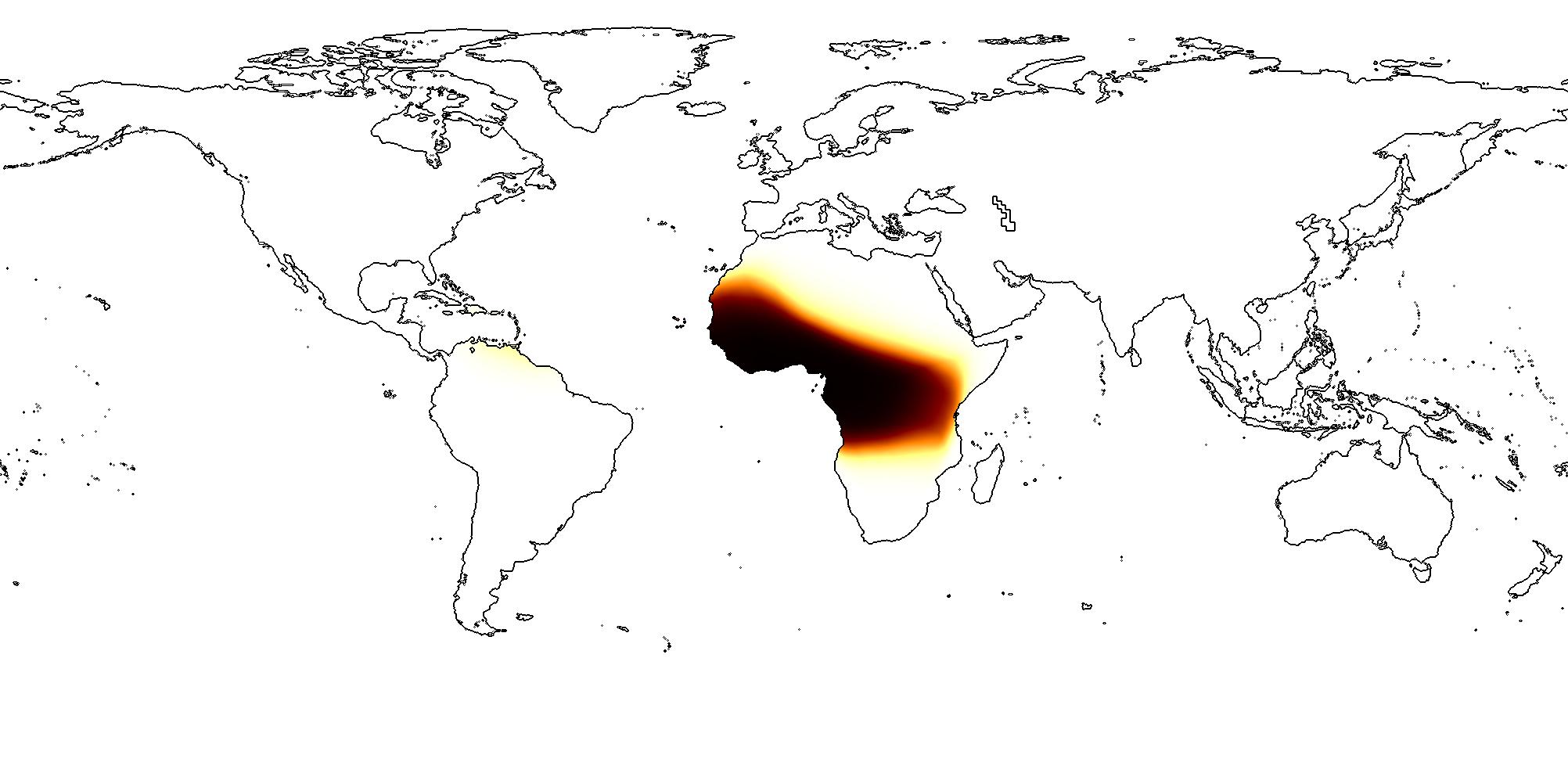}
		\vspacepredd
		\caption[]{{\small 
		$\aodha*$
		}}    
		\label{fig:2891_aodha_exp}
	\end{subfigure}
	\hfill
	\begin{subfigure}[b]{0.24\textwidth}  
		\centering 
		\includegraphics[width=\textwidth]{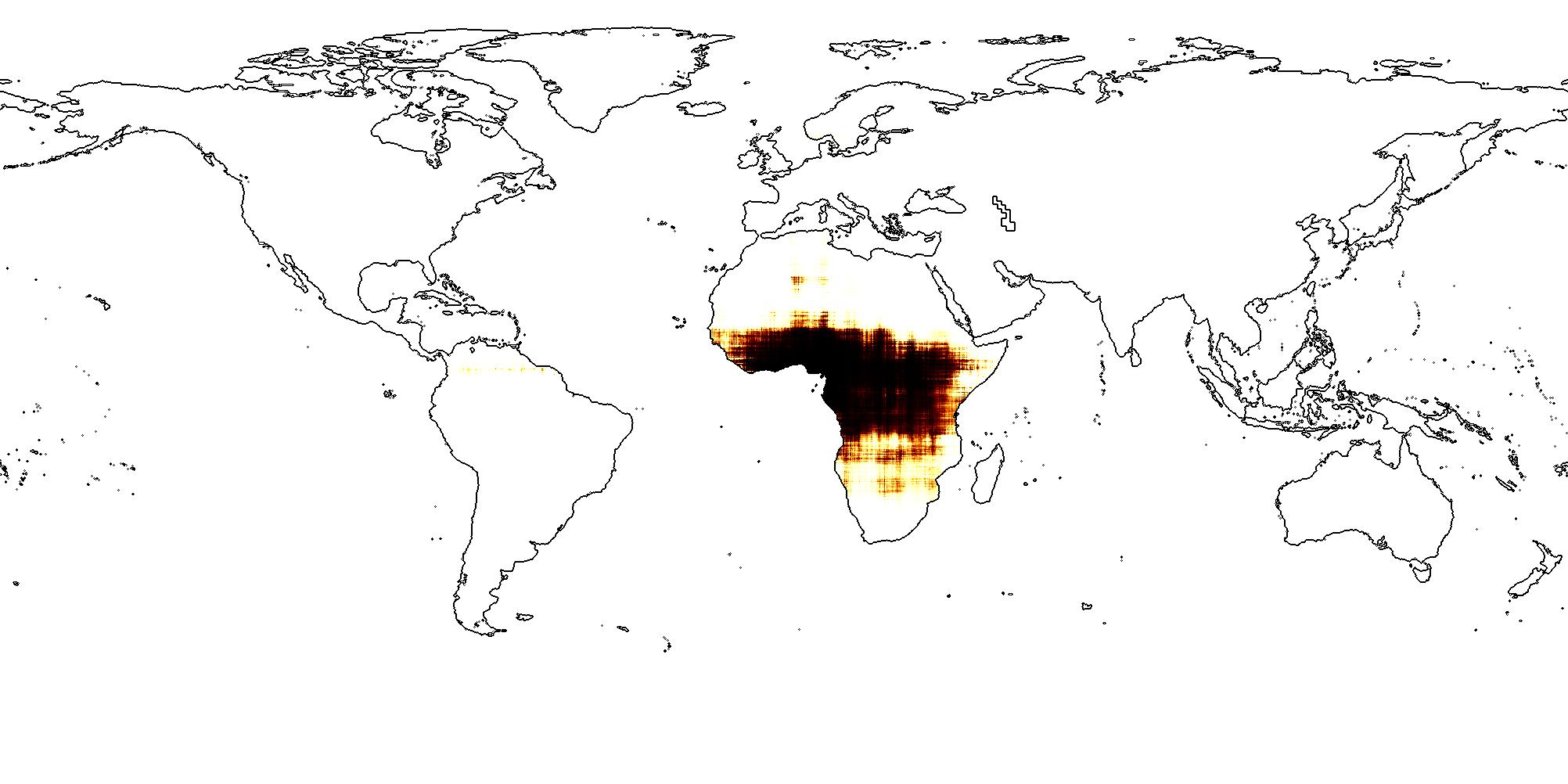}
		\vspacepred
		\caption[]{{\small 
		$\grid$
		}}    
		\label{fig:2891_grid_exp}
	\end{subfigure}
	\hfill
	\begin{subfigure}[b]{0.24\textwidth}  
		\centering 
		\includegraphics[width=\textwidth]{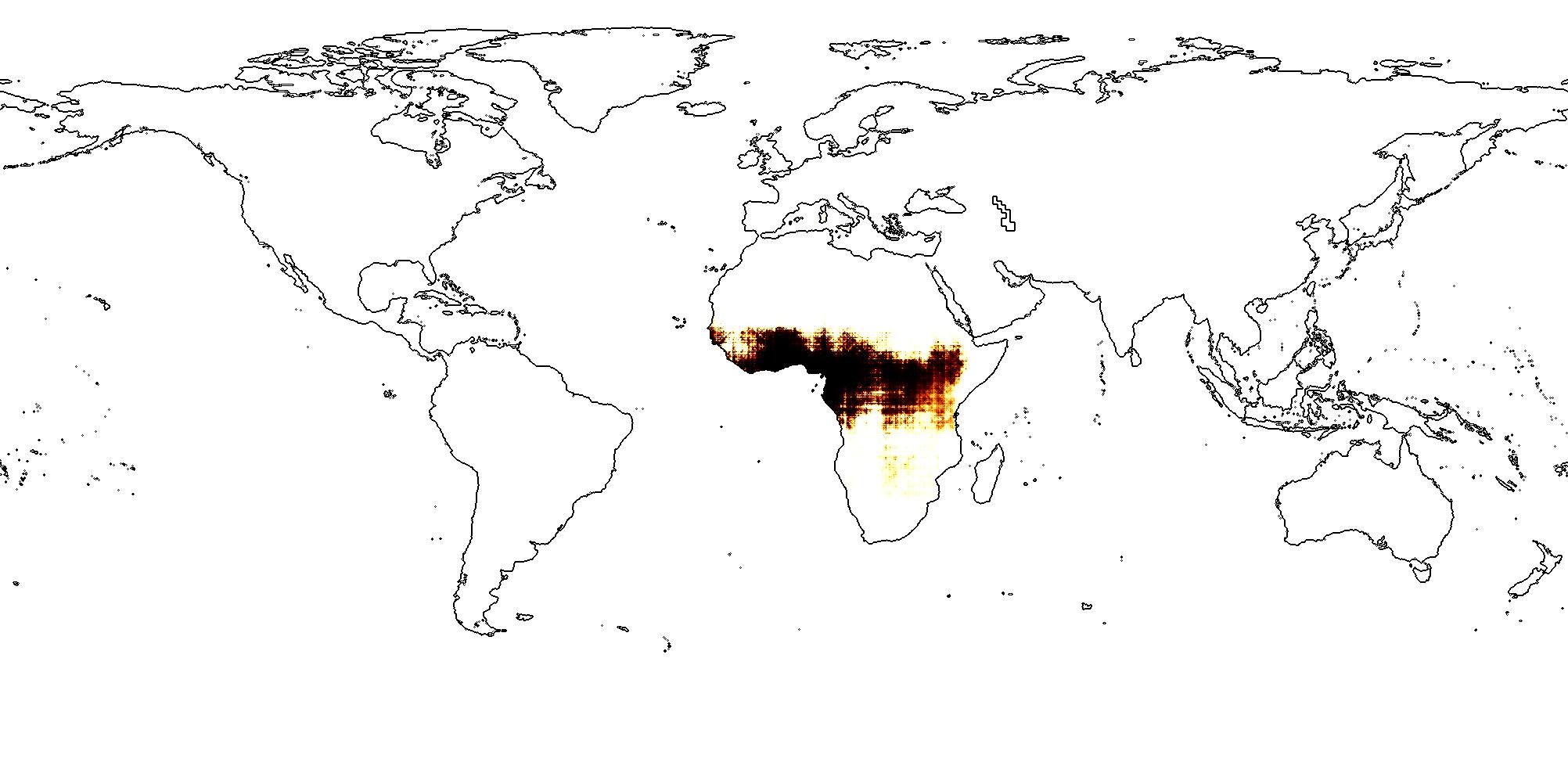}
		\vspacepred
		\caption[]{{\small 
		$\spheregrid$
		}}    
		\label{fig:2891_spheregrid_exp}
	\end{subfigure}

	\caption{Compare the predicted  distributions of example species from different models. 
The first figure of each row marks the data points from iNat2018 training data.
	} 
	\label{fig:spesdist_exp}
	\vspace*{-0.15cm}
\end{figure*}

We further plot the predicted species distributions from different models at different geographic regions, and compare them with the training sample locations of the corresponding species, see Figure \ref{fig:spesdist_intro}b-e and \ref{fig:spesdist_exp}. We can see that compared with $\aodha*$ and $\grid$, in each geographic region with sparse training samples and the North Pole area, the spatial distributions produced by $\spheregrid$ are more compact while the other two have over-generalization issue. For more analysis, please refer to Appendix \ref{sec:sample_impact}, \ref{sec:pred_2018}, and \ref{sec:emb_clustering}.

 \section{Conclusion}  \label{sec:conclusion}
In this work, we propose a general purpose multi-scale spherical location encoder - $\modelname$ which can encode any location on the spherical surface into a high dimension vector which is learning-friendly for downstream neuron network models. We provide theoretical proof that $\modelname$ is able to preserve the spherical surface distance between points. 
We conduct experiments on the geo-aware image classification with 5 large-scale real-world datasets.
Results shows that $\modelname$ can outperform the state-of-the-art 2D location encoders on both tasks. Further analysis shows that $\modelname$ is especially excel at polar regions as well data-sparse areas.

\section{Broader Impact}  \label{sec:impact}
Encoding point-features on a spherical surface is a fundamental problem, especially in geoinformatics, geography, meteorology, oceanography, geoscience, and environmental science. Our proposed $\modelname$ is a general-purpose spherical-distance-reserving encoding which can be utilized in a wide range of geospatial prediction tasks. Except for the tasks we discussed above, the potential applications include areas like public health, epidemiology, agriculture, economy, ecology, and environmental engineering, and researches like large-scale human mobility and trajectory prediction \cite{xu2018encoding}, geographic question answering\cite{mai2020se}, global biodiversity hotspot prediction \cite{myers2000biodiversity,DiMarcoetal2019,Ceballosetal2020}, weather forecasting and climate change \cite{ham2019deep}, global pandemic study and its relation to air pollution \cite{wu2020exposure}, and so on. In general, we expect our proposed $\modelname$ will benefit various \textit{AI for social goods}\footnote{\url{https://ai.google/social-good/}} applications which involves predictive modeling at global scales.

\bibliographystyle{plain}
\bibliography{reference}

\newpage
\section{Appendix}

\subsection{The Map Projection Distortion Problem} \label{sec:map_proj}

\begin{figure*}[t!]
	\centering \tiny
	\vspace*{-0.2cm}
\begin{subfigure}[b]{0.24\textwidth}  
		\centering 
		\includegraphics[width=\textwidth]{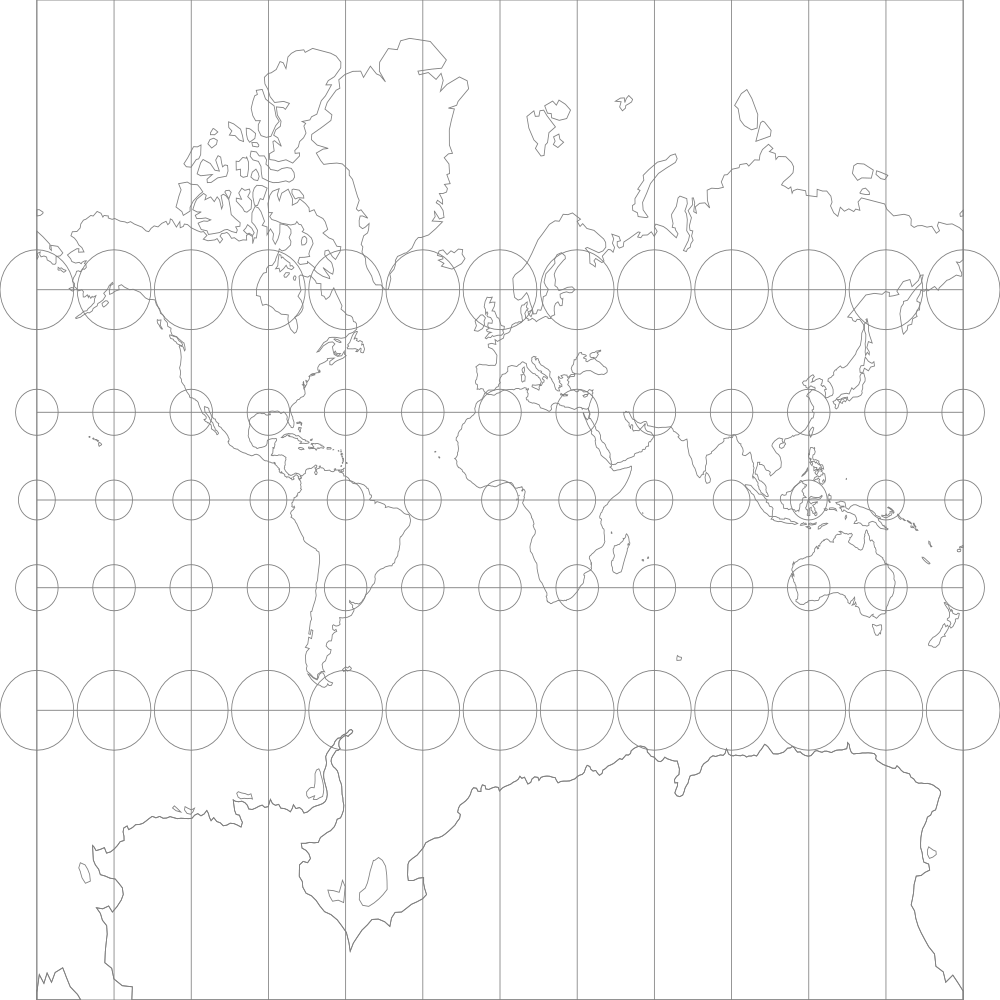}
		\vspace*{-0.3cm}
		\caption[]{{\small 
		Mercator
		}}    
		\label{fig:mercator}
	\end{subfigure}
	\hfill
	\begin{subfigure}[b]{0.24\textwidth}  
		\centering 
		\includegraphics[width=\textwidth]{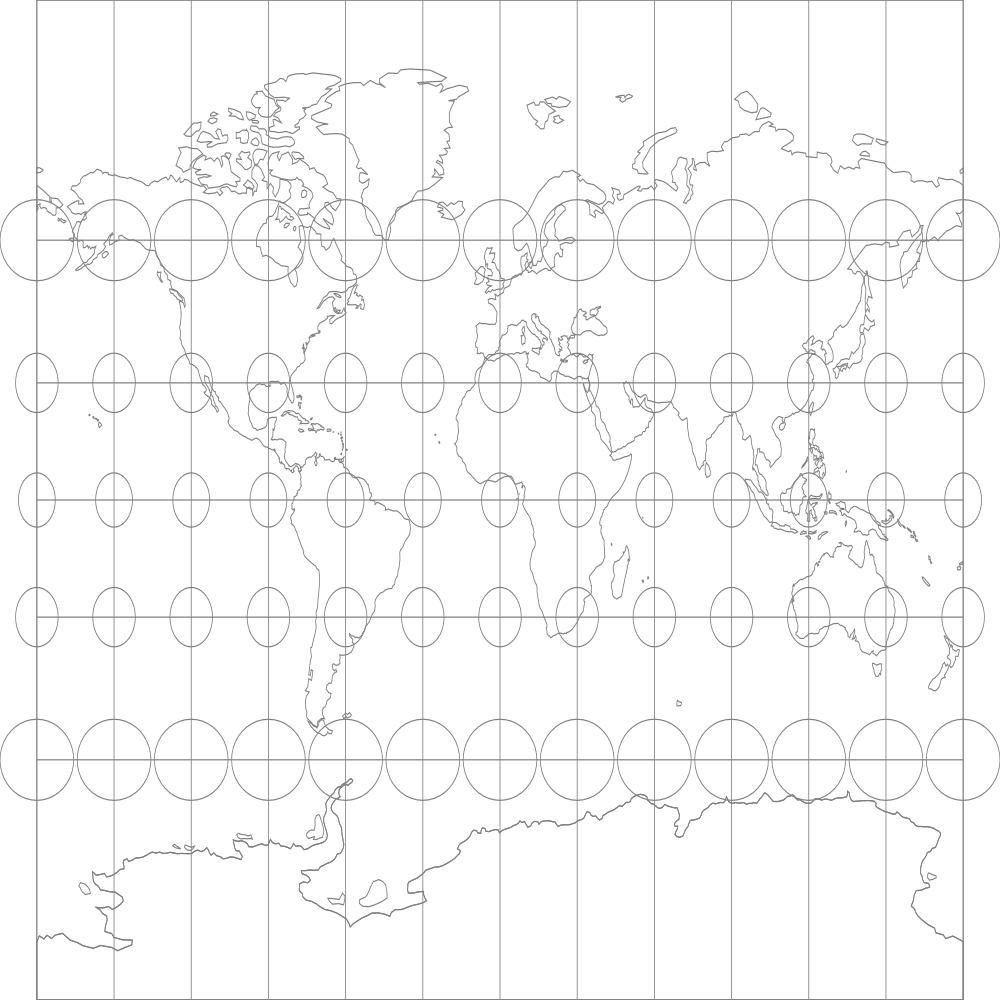}
		\vspace*{-0.3cm}
		\caption[]{{\small 
		Miller
		}}    
		\label{fig:miller}
	\end{subfigure}
	\hfill
	\begin{subfigure}[b]{0.24\textwidth}  
		\centering 
		\includegraphics[width=\textwidth]{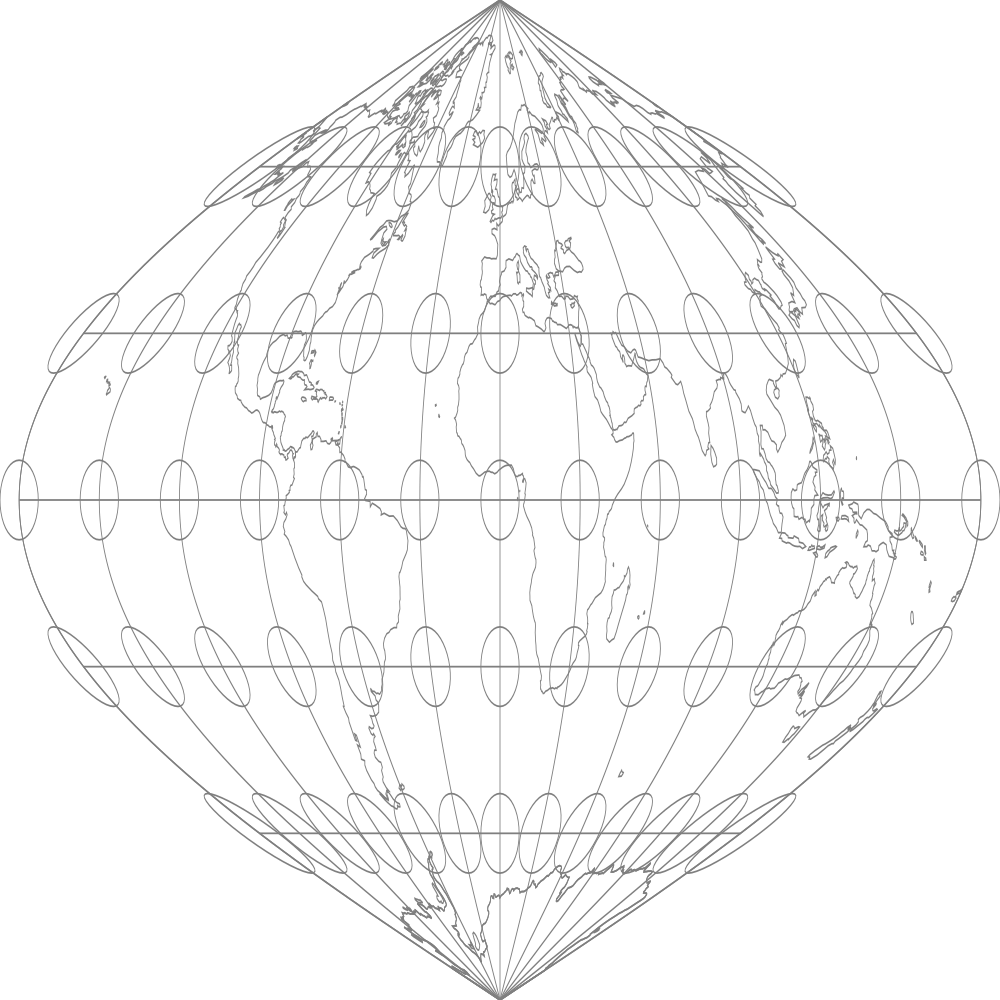}
\caption[]{{\small 
		Sinusoidal
		}}    
		\label{fig:sinusoidal}
	\end{subfigure}
	\hfill
	\begin{subfigure}[b]{0.24\textwidth}  
		\centering 
		\includegraphics[width=\textwidth]{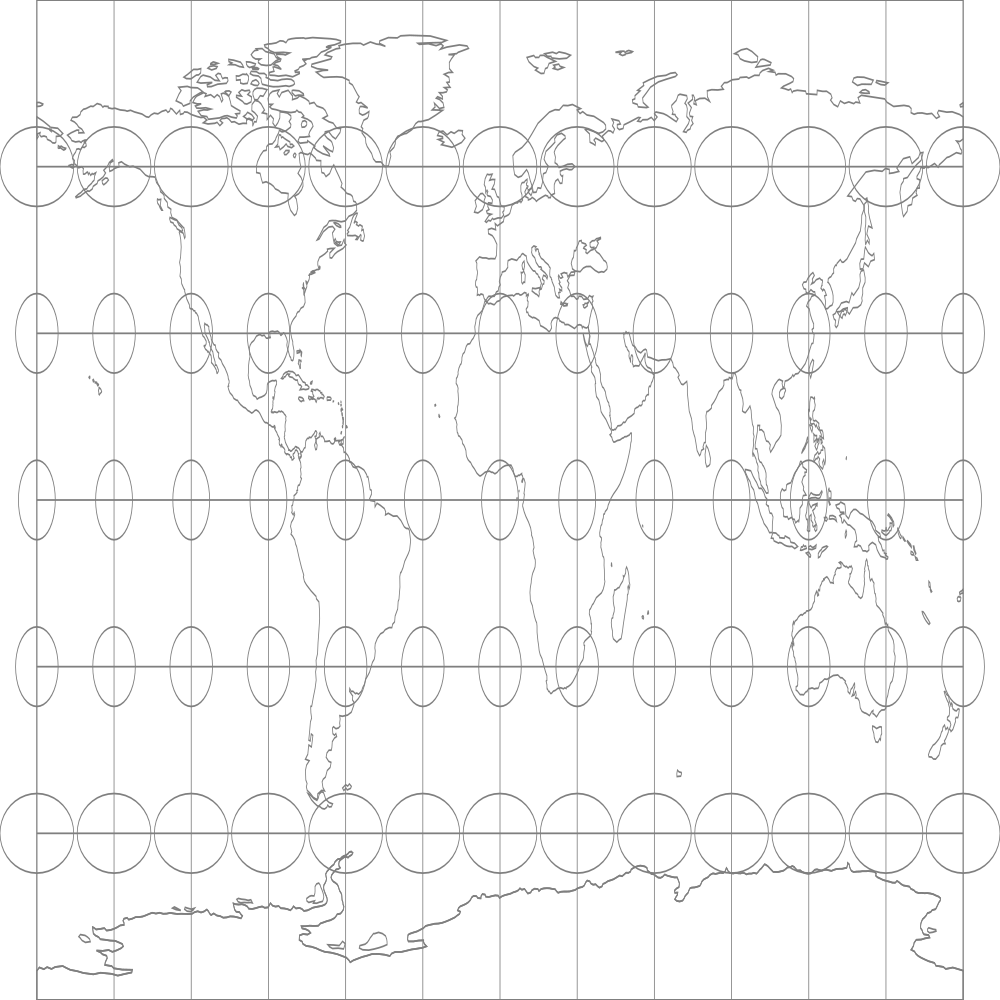}
		\vspace*{-0.3cm}
		\caption[]{{\small 
		Equirectangular
		}}    
		\label{fig:Equirectangle}
	\end{subfigure}
	\caption{An illustration for map projection distortion:
	(a)-(d): Tissot indicatrices for four projections.
The equal area circles are putted in different locations to show how the map distortion affect its shape.
	} 
	\label{fig:map_proj}
	\vspace*{-0.15cm}
\end{figure*}

\textit{Map projection distortion is unavoidable when projecting spherical coordinates into 2D space.} There are no map projection can preserve distances at all direction.
The so-called equidistant projection can only preserve distance on one direction, e.g., the longitude direction for the equirectangular projection (See Figure \ref{fig:Equirectangle}), while the conformal map projections (See Figure \ref{fig:mercator}) can preserve directions while resulting in a large distance distortion. 
For a comprehensive overview of map projections and their distortions, See Mulcahy et al. \cite{mulcahy2001symbolization}. This is a well recognized problem in Cartography which shows the importance of \textit{calculating on a round planet} \cite{chrisman2017calculating}. 

\subsection{A Brief Overview of Geo-Aware Image Classification Task}  \label{subsec:img_cls_loss}
If we assume $\image$ and $\th$ are conditionally independent given $\classy$, then based on Bayes' theorem, we have

\begin{align}
    P(\classy|\image,\th) = \dfrac{P(\image, \th | \classy)P(\classy)}{P(\image, \th)} = \dfrac{P(\image|\classy)P(\image)}{P(\classy)} \dfrac{P(\classy | \th)P(\th)}{P(\classy)} \dfrac{P(\classy)}{P(\image, \th)} \; \propto \; P(\classy|\th)P(\classy|\image)
    \label{eq:img_pred}
\end{align}

Fig \ref{fig:pos_enc} illustrates the whole workflow. The major objective for this task is to learn a geographic prior distribution $P(\classy|\th) \; \propto \; \act(\enc(\th)\classemb_{:,\classy})$ such that all observed species occurrences (all image locations $\th$ as well as their associated species class $\classy$) have maximum probability. Mac Aodha et al. \cite{mac2019presence} used a loss function which is based on maximum likelihood estimation (MLE). Given a set of training samples - data points and their associated class labels $\sampleset = \{(\th, \classy)\}$, the loss function $\imgclsloss(\sampleset)$ is defined as:
\begin{align}
\begin{split}
    \imgclsloss(\sampleset)  = \sum_{(\th, \classy) \in \sampleset} \sum_{\th^{-} \in \negsamp(\th)} \Big ( & \lossweight\log(\act(\enc(\th)\classemb_{:,\classy})) + \\ & \sum_{i=1,i \neq \classy}^{\numclass} \log(1 - \act(\enc(\th)\classemb_{:,i}) ) + 
    \\ & \sum_{i=1}^{\numclass} \log(1 - \act(\enc(\th^{-})\classemb_{:,i}) ) \Big )
\end{split}
\label{equ:imgloss}
\end{align}

Here, $\lossweight$ is a hyperparameter to enlarge the weight of positive samples. $\negsamp(\th)$ represents the negative sample set of point $\th$ while $\th^{-} \in \negsamp(\th)$ is a negative sample uniformly generated from the spherical surface given each data point $\th$. We follow the same model set up but replace the $\aodha$ location encoder with our $\modelname$ model. 

\subsection{ \modelname~ Hyperparameters }\label{sec:param}

Table \ref{tab:imgcls_eval_param} shows the best hyperparameter combinations  of different $\modelname$ models on different image classification dataset. We use a smaller $\freq$ for $\dft$ since it has $O(\freq^{2})$ terms while the other models have $O(\freq)$ terms. $\dft$ with $\freq=8$ yield a similar number of terms to the other models wth $\freq=32$ (See Table \ref{tab:dim}). 
Interestingly, all first four $\modelname$ models ($\sphere$, $\spheregrid$, $\spheremixscale$, and $\spheregridmixscale$) shows the best performance on all five datasets with the same hyperparamter combinations. Note that compared with other datasets, iNat2017 and iNat2018 are more up-to-date datasets with more training samples and better geographic coverage. This indicates that the proposed 4 $\modelname$ models show similar performance over different hyperparameter combinations.

\begin{table}[!]
\caption{The best hyperparameter combinations of $\modelname$ models on different image classification datasets. The learning rate $\lr$ tends to be smaller for larger datasets;
We fix the total number of frequencies $\freq$ to be 8 for $\dft$ and 32 for all others;
the maximum scale $\maxscale=1$;
$\minscale$: the minimum scale;
the number of hidden layers  $\pemlp()$ is fixed to $\numresnet=1$;
the number of neurons in  $\pemlp()$ is fixed to $\numneuron=1024$ except for the smallest dataset.
	}
\label{tab:imgcls_eval_param}
\centering
\begin{tabular}{l|c|c|c}
\toprule
Dataset   &  $\lr$  & $\minscale$ & $\numneuron$ \\ \hline
{BirdSnap} &  0.001  & $10^{-6}$& 512  \\ \hline
{BirdSnap$\dagger$} &  0.001  & $10^{-4}$ & 1024 \\ \hline    
{NABirds$\dagger$ } & 0.001  & $10^{-4}$   & 1024  \\ \hline
{iNat2017} &  0.0001 & $10^{-2}$  & 1024 \\ \hline
{iNat2018} & 0.0005  & $10^{-3}$  & 1024 \\  \bottomrule
\end{tabular}
\end{table}

\begin{table}[]
\centering
\caption{Dimension of position encoding for different models in terms of total scales $S$}
\label{tab:dim}
\begin{tabular}{cccccl}
\hline
Model     & $\sphere$ & $\spheregrid$ & $\spheremixscale$ & $\spheregridmixscale$ & $\dft$   \\ \hline
Dimension & $3S$      & $6S$          & $5S$              & $8S$                  & $4S^2+4S$ \\ \hline
\end{tabular}
\label{modeldim}
\end{table}

\subsection{Impact of MRR by The Number of Samples at Different Latitude Bands}\label{sec:sample_impact}
\begin{figure*}[t!]
	\centering \tiny
	\vspace*{-0.2cm}
	\begin{subfigure}[b]{0.45\textwidth}  
		\centering 
		\includegraphics[width=\textwidth]{./fig/inat_2017_dmrr_num_sample_in_lat.png}\vspace*{-0.2cm}
		\caption[]{{iNat2017
		}}    
		\label{fig:inat2017_mrr_num_sample}
	\end{subfigure}
	\hfill
	\begin{subfigure}[b]{0.45\textwidth}  
		\centering 
		\includegraphics[width=\textwidth]{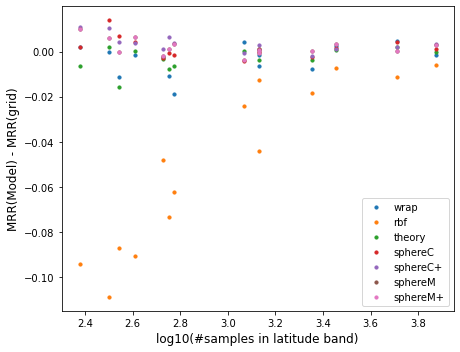}\vspace*{-0.2cm}
		\caption[]{{iNat2018
		}}    
		\label{fig:inat2018_mrr_num_sample}
	\end{subfigure}
	\caption{Impact of MRR by The Number of Samples at Different Latitude Bands. 
	} 
	\label{fig:sample_mrr}
	\vspace*{-0.15cm}
\end{figure*}

See Figure~\ref{fig:sample_mrr}

\subsection{Predicted Distributions iNat2018}\label{sec:pred_2018}
\begin{figure*}[t!]
	\centering \tiny
	\vspace*{-0.2cm}
	\begin{subfigure}[b]{0.24\textwidth}  
		\centering 
		\includegraphics[width=\textwidth]{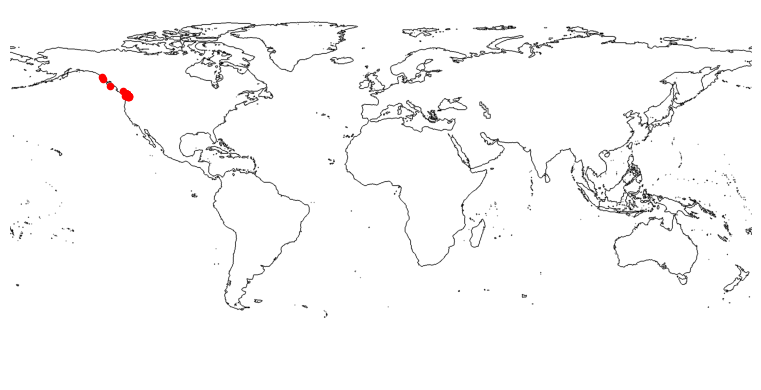}\vspacepred
		\caption[]{{\small 
		Eudistylia vancouveri
		}}    
		\label{fig:0002_dist}
	\end{subfigure}
	\hfill
	\begin{subfigure}[b]{0.24\textwidth}  
		\centering 
		\includegraphics[width=\textwidth]{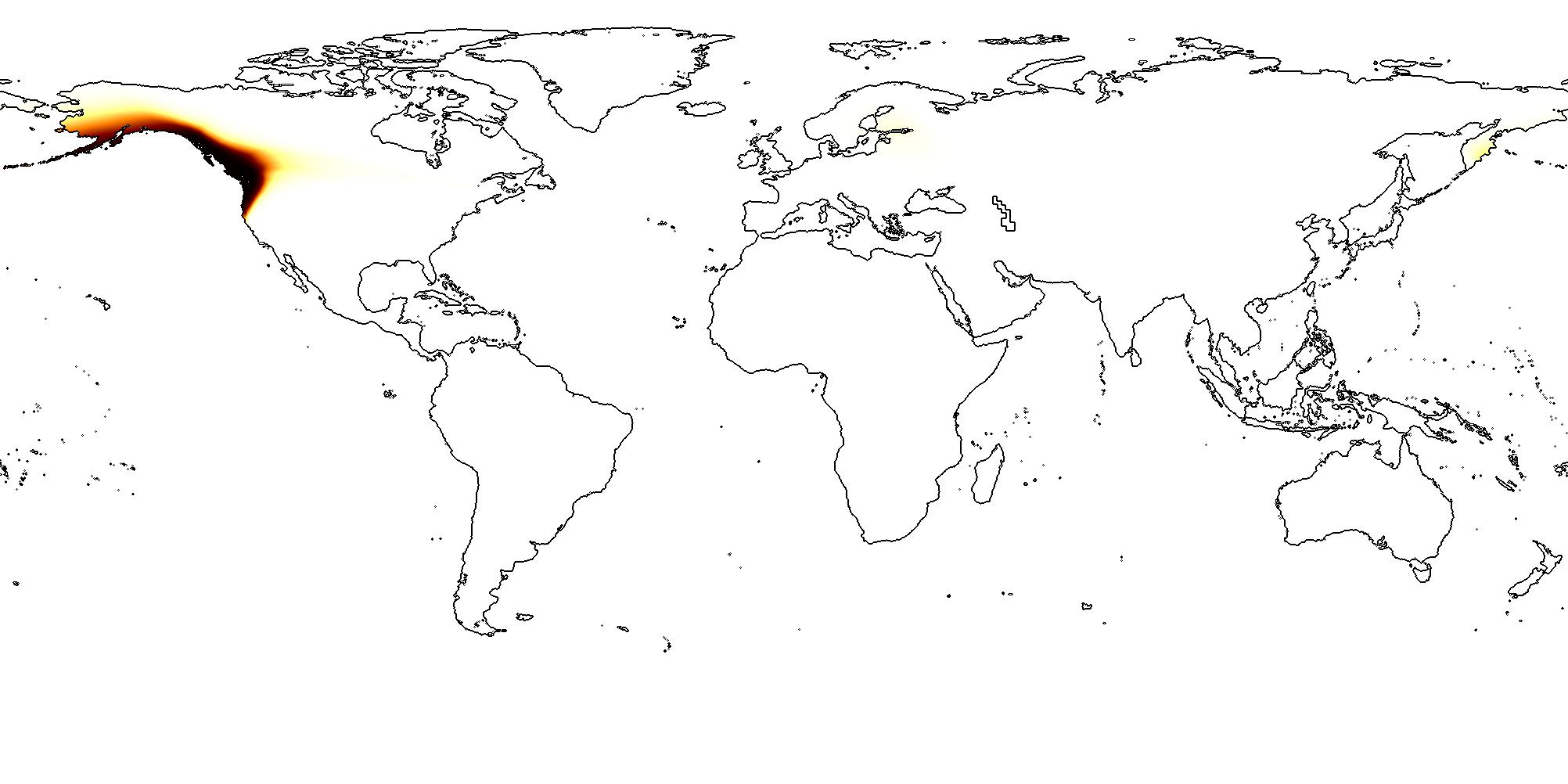}\vspacepred
		\caption[]{{\small 
		$\aodha*$
		}}    
		\label{fig:0002_aodha}
	\end{subfigure}
	\hfill
	\begin{subfigure}[b]{0.24\textwidth}  
		\centering 
		\includegraphics[width=\textwidth]{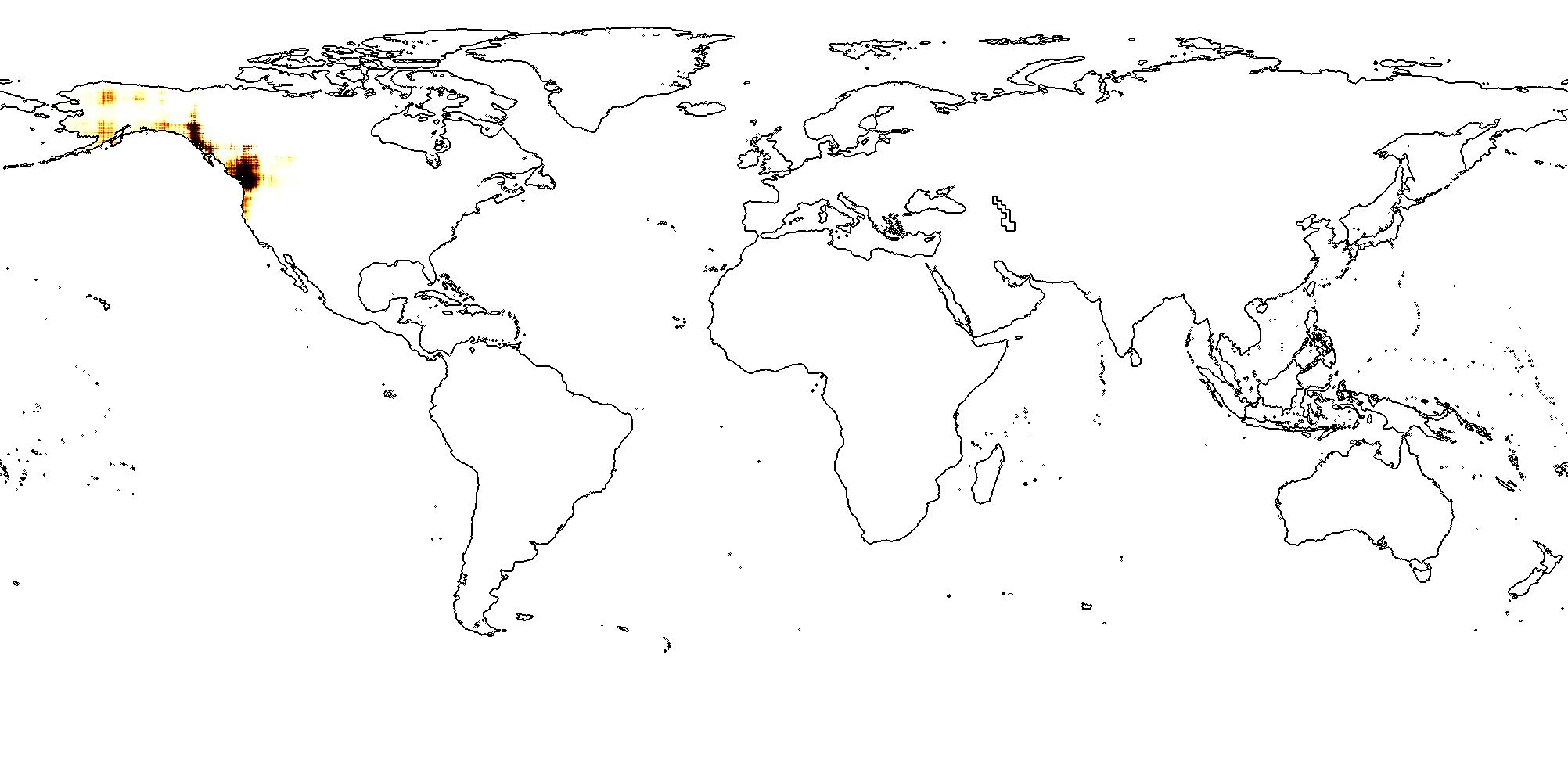}\vspacepred
		\caption[]{{\small 
		$\grid$
		}}    
		\label{fig:0002_grid}
	\end{subfigure}
	\hfill
	\begin{subfigure}[b]{0.24\textwidth}  
		\centering 
		\includegraphics[width=\textwidth]{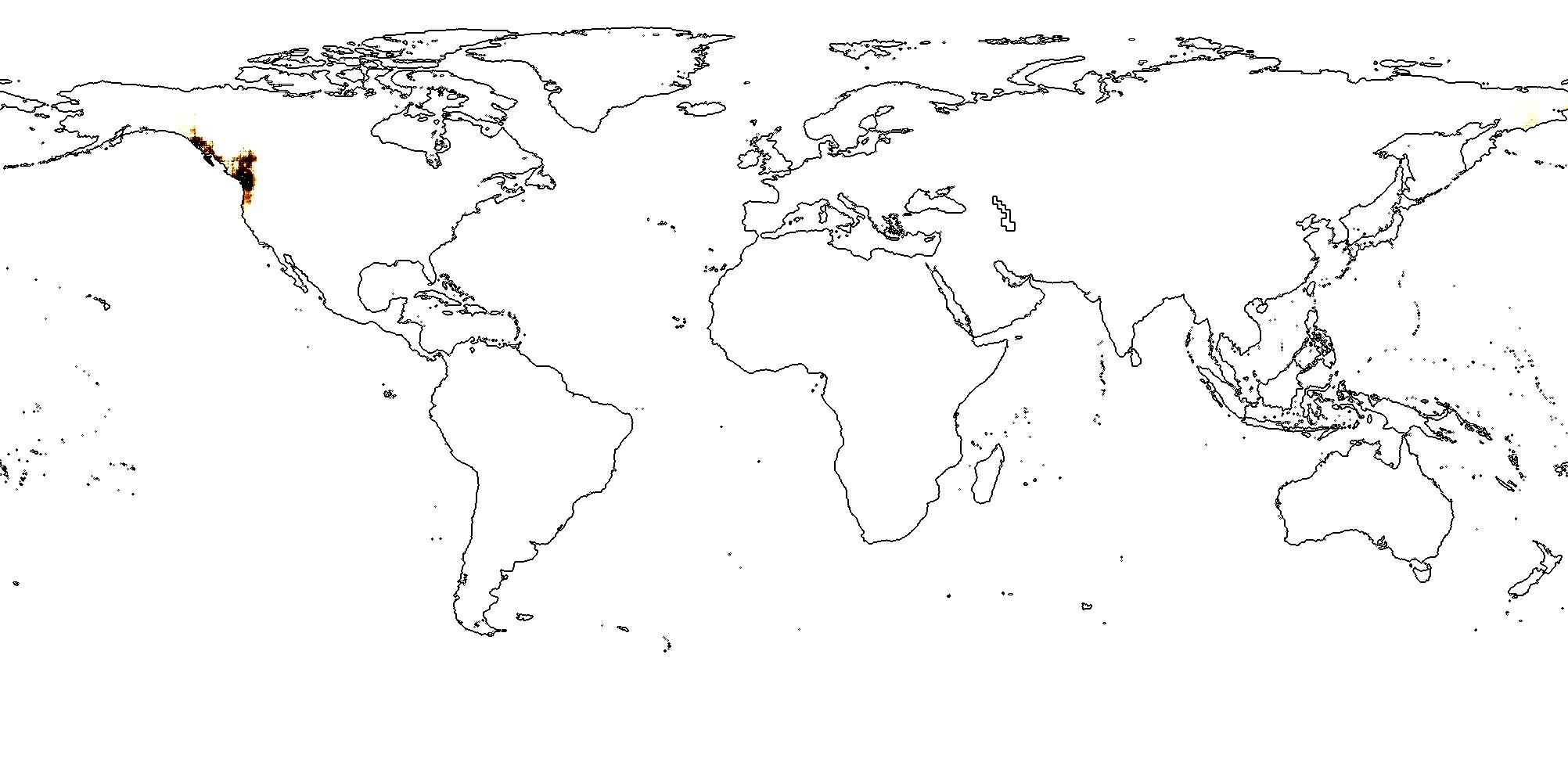}\vspacepred
		\caption[]{{\small 
		$\spheregrid$
		}}    
		\label{fig:0002_spheregrid}
	\end{subfigure}
	\hfill
	\begin{subfigure}[b]{0.24\textwidth}  
		\centering 
		\includegraphics[width=\textwidth]{./fig/3475_cnt_27.png}\vspacepred
		\caption[]{{\scriptsize 
		Motacilla maderaspatensis
		}}    
		\label{fig:3475_dist}
	\end{subfigure}
	\hfill
	\begin{subfigure}[b]{0.24\textwidth}  
		\centering 
		\includegraphics[width=\textwidth]{./fig/gt_3475_Motacilla-maderaspatensis_geo_net_predict.jpg}\vspacepred
		\caption[]{{\small 
		$\aodha*$
		}}    
		\label{fig:3475_aodha}
	\end{subfigure}
	\hfill
	\begin{subfigure}[b]{0.24\textwidth}  
		\centering 
		\includegraphics[width=\textwidth]{./fig/gt_3475_Motacilla-maderaspatensis_gridcell_predict.jpg}\vspacepred
		\caption[]{{\small 
		$\grid$
		}}    
		\label{fig:3475_grid}
	\end{subfigure}
	\hfill
	\begin{subfigure}[b]{0.24\textwidth}  
		\centering 
		\includegraphics[width=\textwidth]{./fig/gt_3475_Motacilla-maderaspatensis_spheregrid_predict.jpg}\vspacepred
		\caption[]{{\small 
		$\spheregrid$
		}}    
		\label{fig:3475_spheregrid}
	\end{subfigure}
	\hfill
	\begin{subfigure}[b]{0.24\textwidth}  
		\centering 
		\includegraphics[width=\textwidth]{./fig/4084_cnt_25.png}\vspacepred
		\caption[]{{\small 
		Vulpes lagopus
		}}    
		\label{fig:4084_dist}
	\end{subfigure}
	\hfill
	\begin{subfigure}[b]{0.24\textwidth}  
		\centering 
		\includegraphics[width=\textwidth]{./fig/gt_4084_Vulpes-lagopus_geo_net_predict.jpg}\vspacepred
		\caption[]{{\small 
		$\aodha*$
		}}    
		\label{fig:4084_aodha}
	\end{subfigure}
	\hfill
	\begin{subfigure}[b]{0.24\textwidth}  
		\centering 
		\includegraphics[width=\textwidth]{./fig/gt_4084_Vulpes-lagopus_gridcell_predict.jpg}\vspacepred
		\caption[]{{\small 
		$\grid$
		}}    
		\label{fig:4084_grid}
	\end{subfigure}
	\hfill
	\begin{subfigure}[b]{0.24\textwidth}  
		\centering 
		\includegraphics[width=\textwidth]{./fig/gt_4084_Vulpes-lagopus_spheregrid_predict.jpg}\vspacepred
		\caption[]{{\small 
		$\spheregrid$
		}}    
		\label{fig:4084_spheregrid}
	\end{subfigure}
	\hfill
	\begin{subfigure}[b]{0.24\textwidth}  
		\centering 
		\includegraphics[width=\textwidth]{./fig/1555_cnt_25.png}\vspacepred
		\caption[]{{\small 
		Siderone galanthis
		}}    
		\label{fig:1555_dist}
	\end{subfigure}
	\hfill
	\begin{subfigure}[b]{0.24\textwidth}  
		\centering 
		\includegraphics[width=\textwidth]{./fig/gt_1555_Siderone-galanthis_geo_net_predict.jpg}\vspacepred
		\caption[]{{\small 
		$\aodha*$
		}}    
		\label{fig:1555_aodha}
	\end{subfigure}
	\hfill
	\begin{subfigure}[b]{0.24\textwidth}  
		\centering 
		\includegraphics[width=\textwidth]{./fig/gt_1555_Siderone-galanthis_gridcell_predict.jpg}\vspacepred
		\caption[]{{\small 
		$\grid$
		}}    
		\label{fig:1555_grid}
	\end{subfigure}
	\hfill
	\begin{subfigure}[b]{0.24\textwidth}  
		\centering 
		\includegraphics[width=\textwidth]{./fig/gt_1555_Siderone-galanthis_spheregrid_predict.jpg}\vspacepred
		\caption[]{{\small 
		$\spheregrid$
		}}    
		\label{fig:1555_spheregrid}
	\end{subfigure}
	\hfill
	\begin{subfigure}[b]{0.24\textwidth}  
		\centering 
		\includegraphics[width=\textwidth]{./fig/2891_cnt_25.png}\vspacepred
		\caption[]{{\small 
		Lophoceros fasciatus
		}}    
		\label{fig:2891_dist}
	\end{subfigure}
	\hfill
	\begin{subfigure}[b]{0.24\textwidth}  
		\centering 
		\includegraphics[width=\textwidth]{./fig/gt_2891_Lophoceros-fasciatus_geo_net_predict.jpg}\vspacepred
		\caption[]{{\small 
		$\aodha*$
		}}    
		\label{fig:2891_aodha}
	\end{subfigure}
	\hfill
	\begin{subfigure}[b]{0.24\textwidth}  
		\centering 
		\includegraphics[width=\textwidth]{./fig/gt_2891_Lophoceros-fasciatus_gridcell_predict.jpg}\vspacepred
		\caption[]{{\small 
		$\grid$
		}}    
		\label{fig:2891_grid}
	\end{subfigure}
	\hfill
	\begin{subfigure}[b]{0.24\textwidth}  
		\centering 
		\includegraphics[width=\textwidth]{./fig/gt_2891_Lophoceros-fasciatus_spheregrid_predict.jpg}\vspacepred
		\caption[]{{\small 
		$\spheregrid$
		}}    
		\label{fig:2891_spheregrid}
	\end{subfigure}
	\hfill
	\begin{subfigure}[b]{0.24\textwidth}  
		\centering 
		\includegraphics[width=\textwidth]{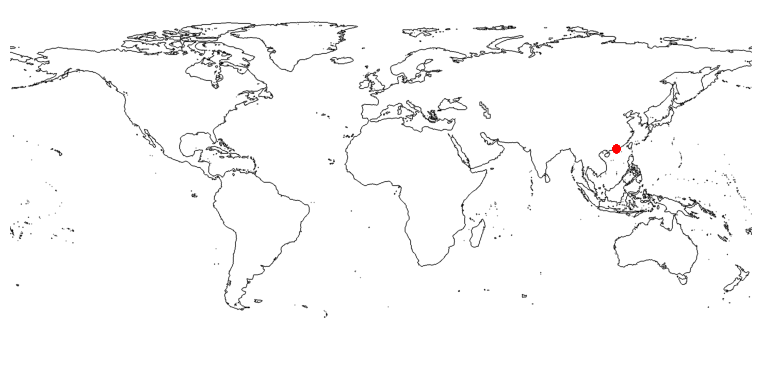}\vspacepred
		\caption[]{{\small 
		Dysphania militaris
		}}    
		\label{fig:0904_dist}
	\end{subfigure}
	\hfill
	\begin{subfigure}[b]{0.24\textwidth}  
		\centering 
		\includegraphics[width=\textwidth]{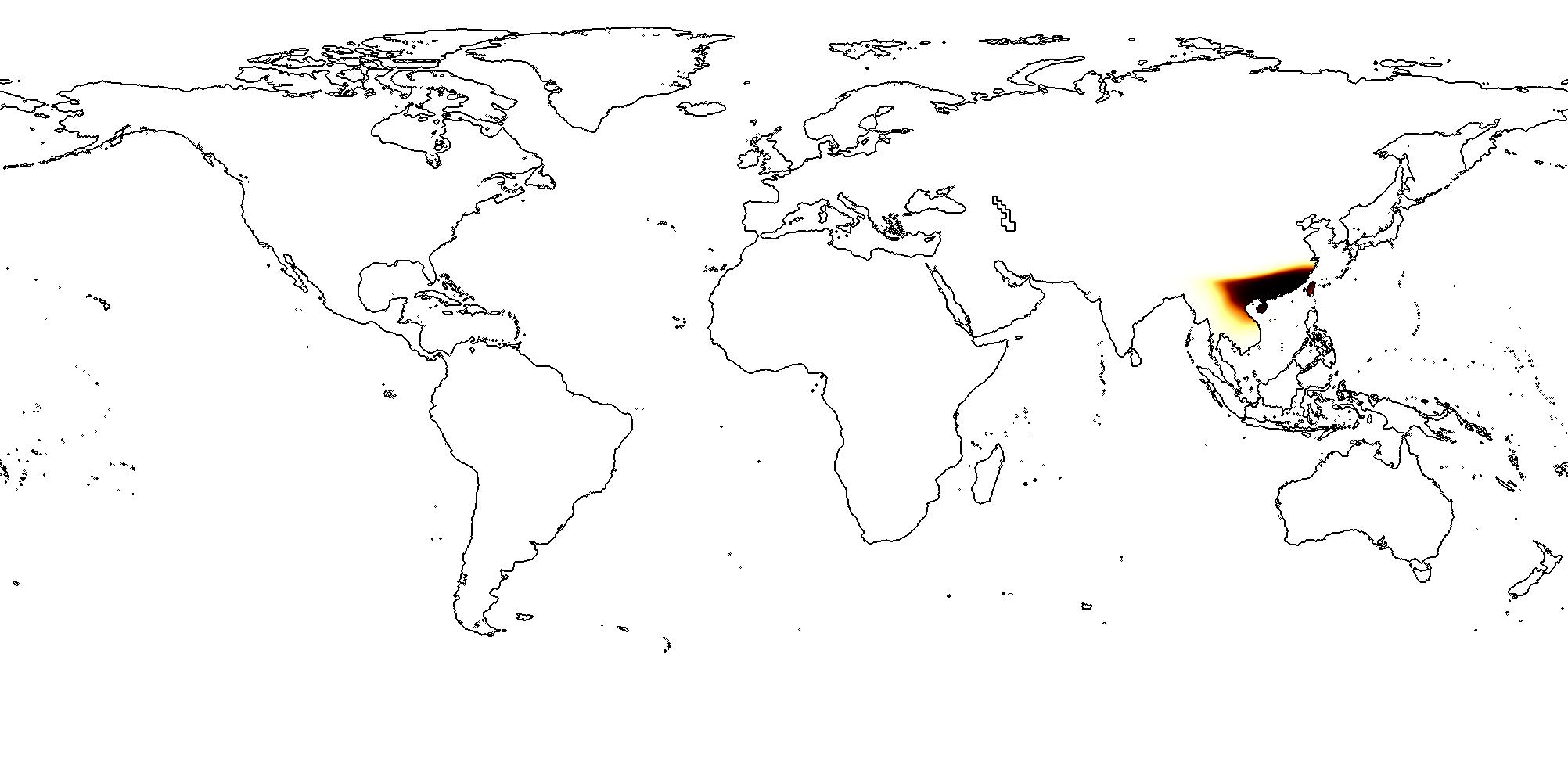}\vspacepred
		\caption[]{{\small 
		$\aodha*$
		}}    
		\label{fig:0904_aodha}
	\end{subfigure}
	\hfill
	\begin{subfigure}[b]{0.24\textwidth}  
		\centering 
		\includegraphics[width=\textwidth]{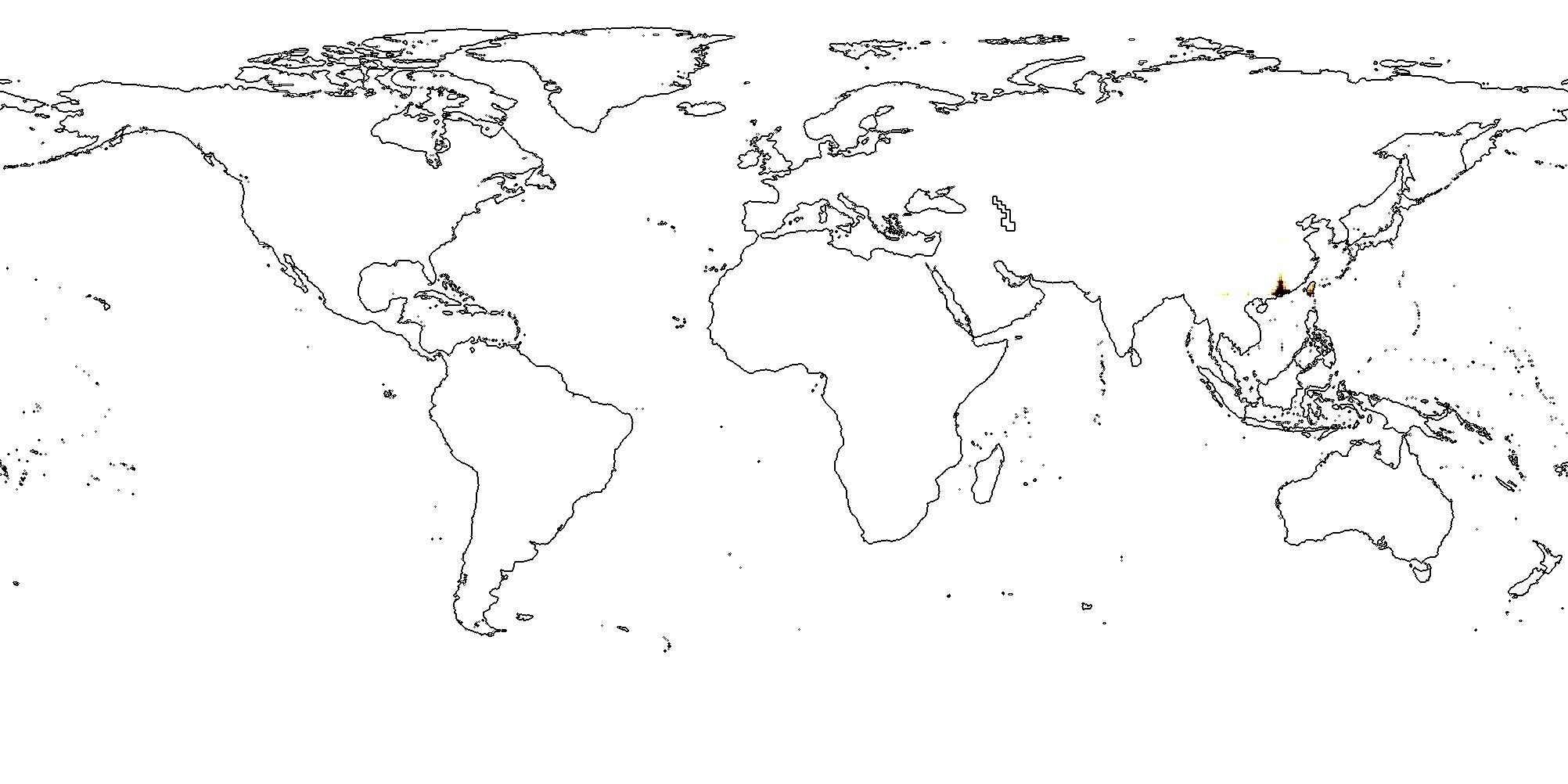}\vspacepred
		\caption[]{{\small 
		$\grid$
		}}    
		\label{fig:0904_grid}
	\end{subfigure}
	\hfill
	\begin{subfigure}[b]{0.24\textwidth}  
		\centering 
		\includegraphics[width=\textwidth]{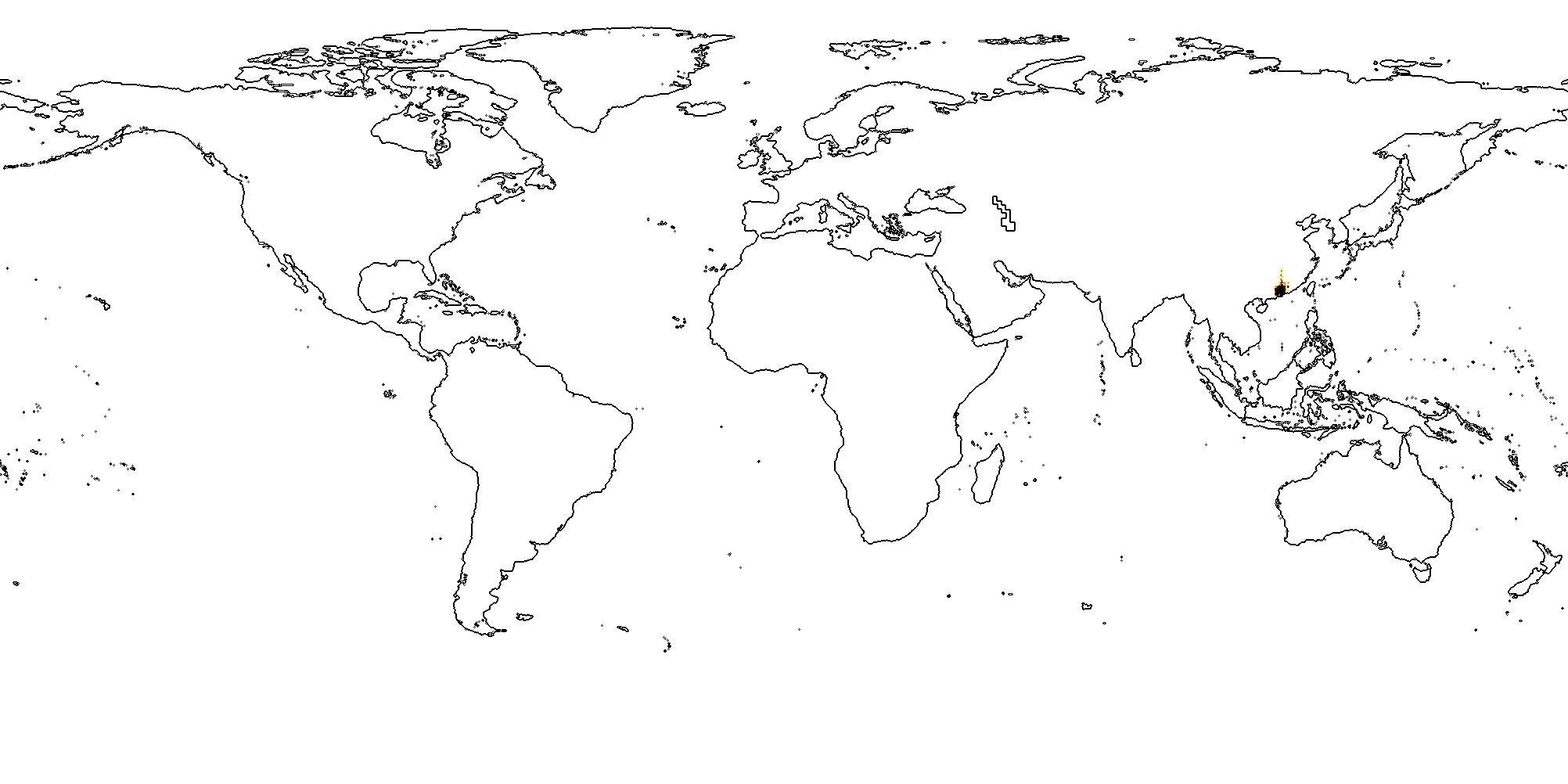}\vspacepred
		\caption[]{{\small 
		$\spheregrid$
		}}    
		\label{fig:0904_spheregrid}
	\end{subfigure}
	
	\caption{Compare the predicted  distributions of example species from different models. 
The first figure of each row marks the data points from iNat2018 training data.
	} 
	\label{fig:spesdist18}
	\vspace*{-0.15cm}
\end{figure*}

See Figure~\ref{fig:spesdist18}

\subsection{Embedding Clustering}\label{sec:emb_clustering}
\begin{figure*}[t!]
	\centering \tiny

	\vspace*{-0.2cm}
	\begin{subfigure}[b]{0.33\textwidth}  
		\centering 
		\includegraphics[width=\textwidth]{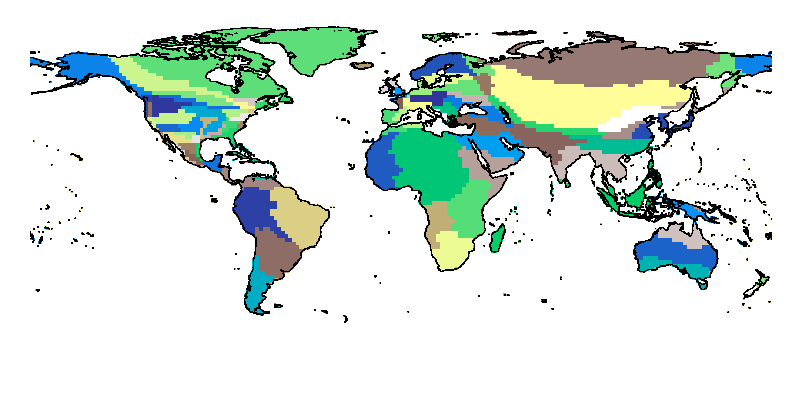}\vspaceclustering
		\caption[]{{\small 
		$\aodha*$ }}    
		\label{fig:inat17_aodha}
	\end{subfigure}
	\hfill
	\begin{subfigure}[b]{0.33\textwidth}  
		\centering 
		\includegraphics[width=\textwidth]{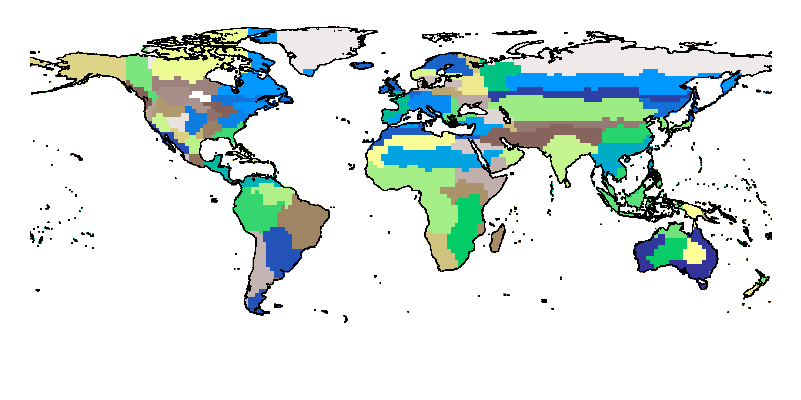}\vspaceclustering
		\caption[]{{\small 
		$\grid$ $(\minscale = 10^{-2})$
		}}    
		\label{fig:inat17_grid_2}
	\end{subfigure}
	\hfill
	\begin{subfigure}[b]{0.33\textwidth}  
		\centering 
		\includegraphics[width=\textwidth]{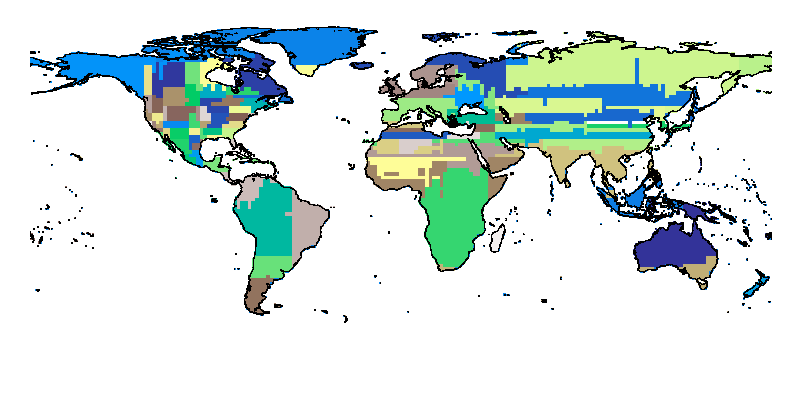}\vspaceclustering
		\caption[]{{\small 
		$\grid$ $(\minscale = 10^{-6})$
		}}    
		\label{fig:inat17_grid}
	\end{subfigure}

	\hfill
	\begin{subfigure}[b]{0.33\textwidth}  
		\centering 
		\includegraphics[width=\textwidth]{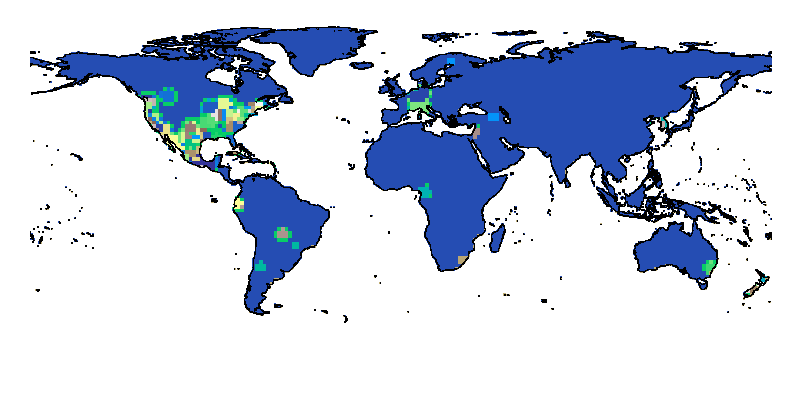}\vspaceclustering
		\caption[]{{\small 
		$\rbf$ $(\sigma = 1, \numkernel = 200)$
		}}    
		\label{fig:inat17_rbf}
	\end{subfigure}
	\hfill
	\begin{subfigure}[b]{0.33\textwidth}  
		\centering 
		\includegraphics[width=\textwidth]{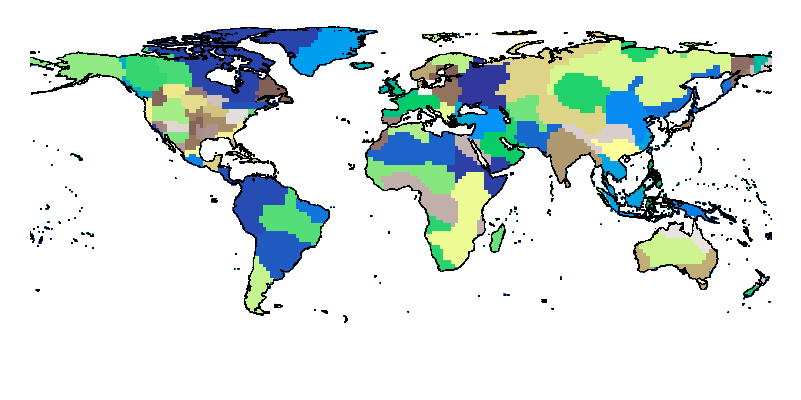}\vspaceclustering
		\caption[]{{\small 
		$\theory$ $ (\minscale = 10^{-2})$
		}}    
		\label{fig:inat17_theory_2}
	\end{subfigure}
	\hfill
	\begin{subfigure}[b]{0.33\textwidth}  
		\centering 
		\includegraphics[width=\textwidth]{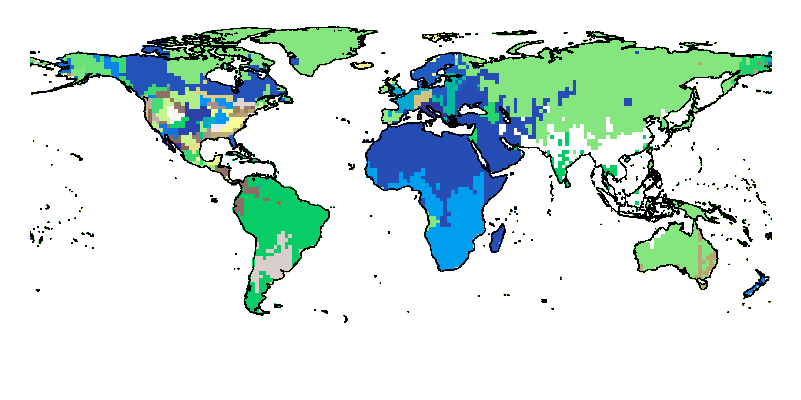}\vspaceclustering
		\caption[]{{\small 
		$\theory$ $ (\minscale = 10^{-6})$
		}}    
		\label{fig:inat17_theory}
	\end{subfigure}
	\hfill
	\begin{subfigure}[b]{0.33\textwidth}  
		\centering 
		\includegraphics[width=\textwidth]{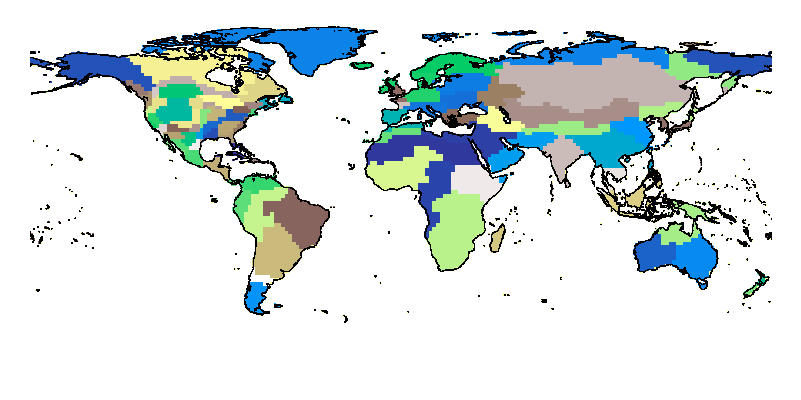}\vspaceclustering
		\caption[]{{\small 
		$\spheremixscale$ $ (\minscale = 10^{-1})$
		}}    
		\label{fig:inat17_spheremixscale_1}
	\end{subfigure}
	\hfill
	\begin{subfigure}[b]{0.33\textwidth}  
		\centering 
		\includegraphics[width=\textwidth]{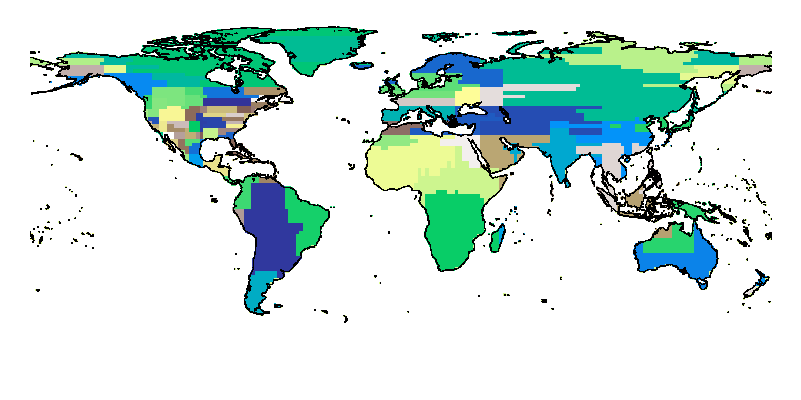}\vspaceclustering
		\caption[]{{\small 
		$\spheremixscale$ $ (\minscale = 10^{-2})$
		}}    
		\label{fig:inat17_spheremixscale_2}
	\end{subfigure}
\hfill
	\begin{subfigure}[b]{0.33\textwidth}  
		\centering 
		\includegraphics[width=\textwidth]{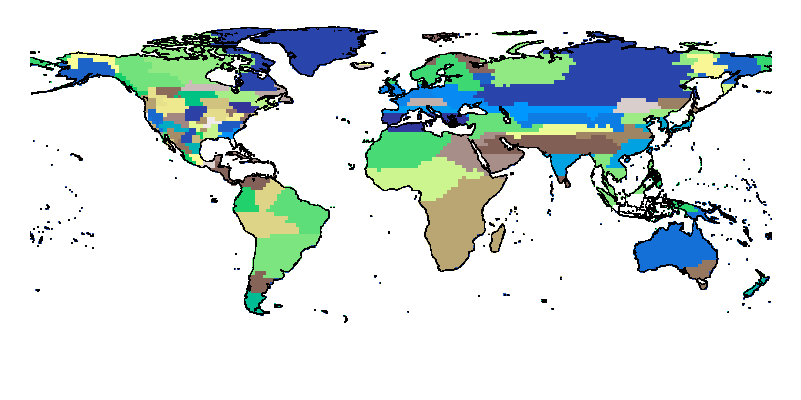}\vspaceclustering
		\caption[]{{\small 
		$\sphere$ $ (\minscale = 10^{-2})$
		}}    
		\label{fig:inat17_sphere}
	\end{subfigure}
	\hfill
	\begin{subfigure}[b]{0.33\textwidth}  
		\centering 
		\includegraphics[width=\textwidth]{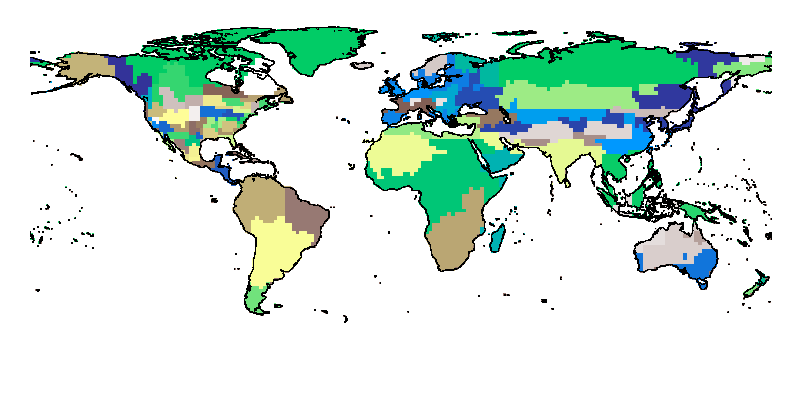}\vspaceclustering
		\caption[]{{\small 
		$\spheregrid$ $ (\minscale = 10^{-2})$
		}}    
		\label{fig:inat17_spheregrid}
	\end{subfigure}
	\hfill
	\begin{subfigure}[b]{0.33\textwidth}  
		\centering 
		\includegraphics[width=\textwidth]{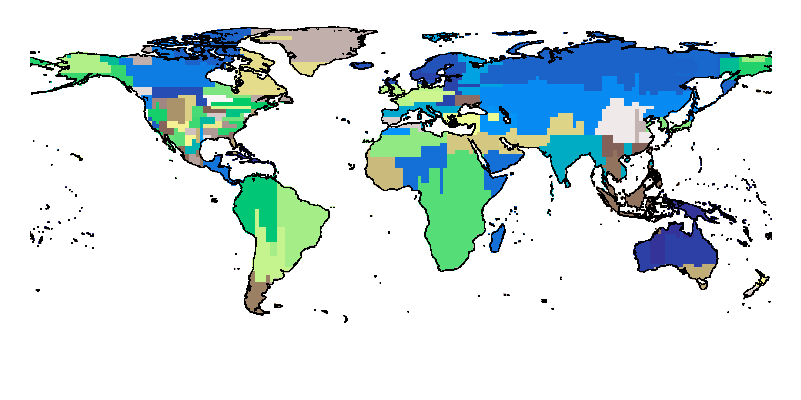}\vspaceclustering
		\caption[]{{\small 
		$\spheregridmixscale$ $ (\minscale = 10^{-2})$
		}}    
		\label{fig:inat17_spheregridmixscale}
	\end{subfigure}
	\hfill
	\begin{subfigure}[b]{0.33\textwidth}  
		\centering 
		\includegraphics[width=\textwidth]{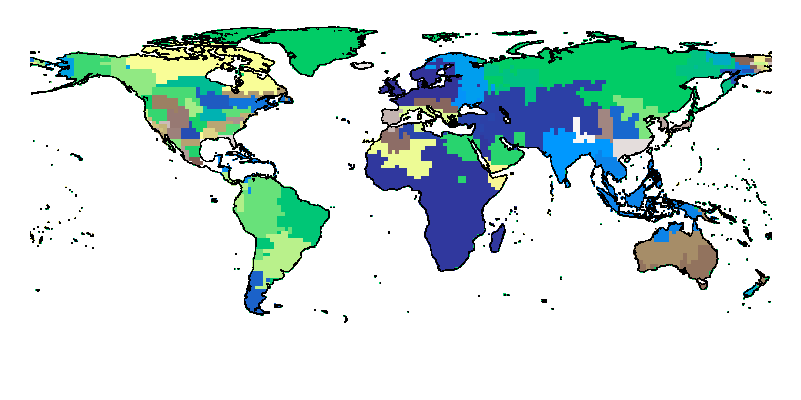}\vspaceclustering
		\caption[]{{\small 
		$\dft$ $ (\minscale = 10^{-2})$
		}}    
		\label{fig:inat17_dft}
	\end{subfigure}

	\caption[]{Embedding clusterings of iNat2017 models. 
	(a) $\aodha*$ with 4 hidden ReLU layers of 256 neurons; 
	(d) $\rbf$ with the best kernel size $\kernelsize=1$ and number of anchor points $\numkernel=200$; 
(b)(c)(e)(f) are $\spacevec$ models \cite{mai2020multiscale} with different min scale $\minscale = \{10^{-6}, 10^{-2}\}$.\textsuperscript{a}
(g)-(l) are different $\modelname$ models.\textsuperscript{b}
} 
	\scriptsize
	\textsuperscript{a} {They share the same best hyperparameters:  $\freq = 64$,  $\maxscale = 1$, and 1 hidden ReLU layers of 512 neurons.} \\
	\textsuperscript{b}{They share the same best hyperparameters: $\freq = 32$, $\maxscale = 1$, and 1 hidden ReLU layers of 1024 neurons.}
	\label{fig:inat17}
\end{figure*}

\begin{figure*}[t!]
	\centering \tiny
	\vspace*{-0.2cm}
	\begin{subfigure}[b]{0.33\textwidth}  
		\centering 
		\includegraphics[width=\textwidth]{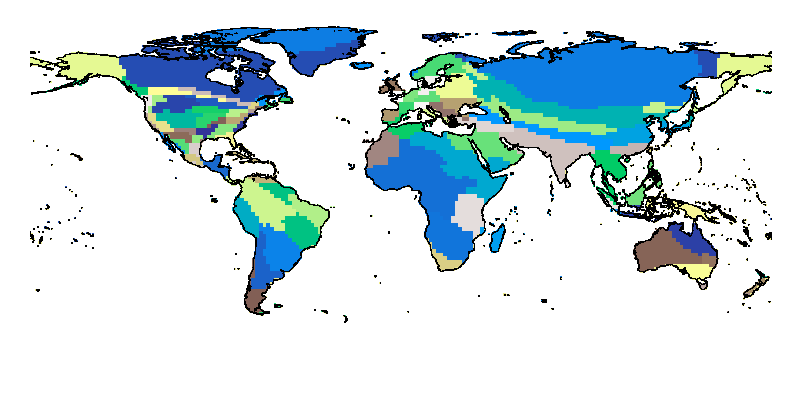}\vspaceclustering
		\caption[]{{\small 
		$\aodha*$ }}    
		\label{fig:inat18_aodha}
	\end{subfigure}
	\hfill
	\begin{subfigure}[b]{0.33\textwidth}  
		\centering 
		\includegraphics[width=\textwidth]{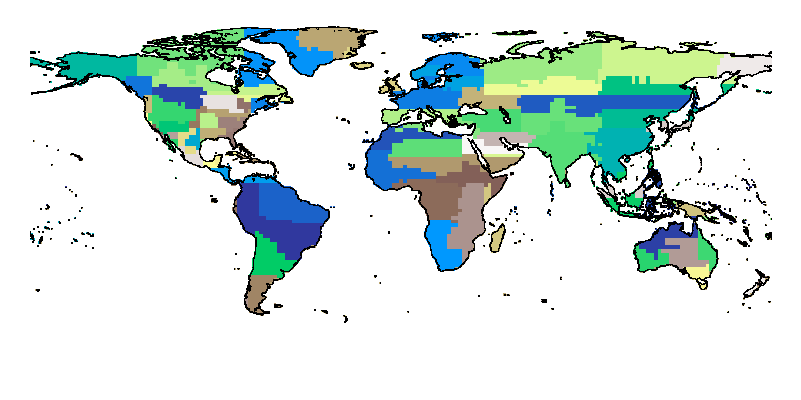}\vspaceclustering
		\caption[]{{\small 
		$\grid$ $(\minscale = 10^{-3})$
		}}    
		\label{fig:inat18_grid_3}
	\end{subfigure}
	\hfill
	\begin{subfigure}[b]{0.33\textwidth}  
		\centering 
		\includegraphics[width=\textwidth]{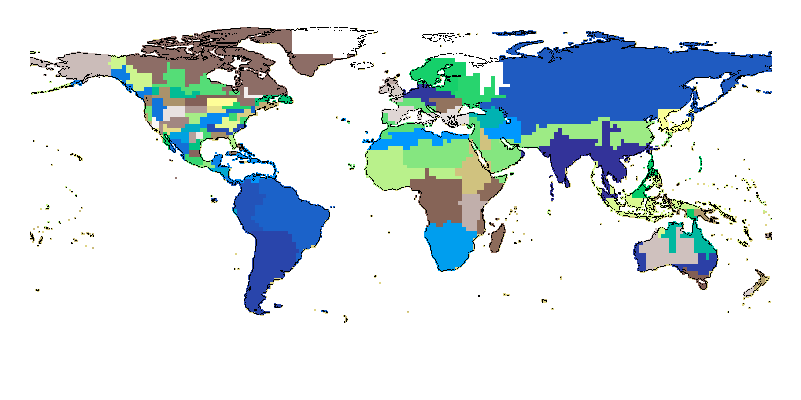}\vspaceclustering
		\caption[]{{\small 
		$\grid$ $(\minscale = 10^{-6})$
		}}    
		\label{fig:inat18_grid}
	\end{subfigure}
	\hfill
	\begin{subfigure}[b]{0.33\textwidth}  
		\centering 
		\includegraphics[width=\textwidth]{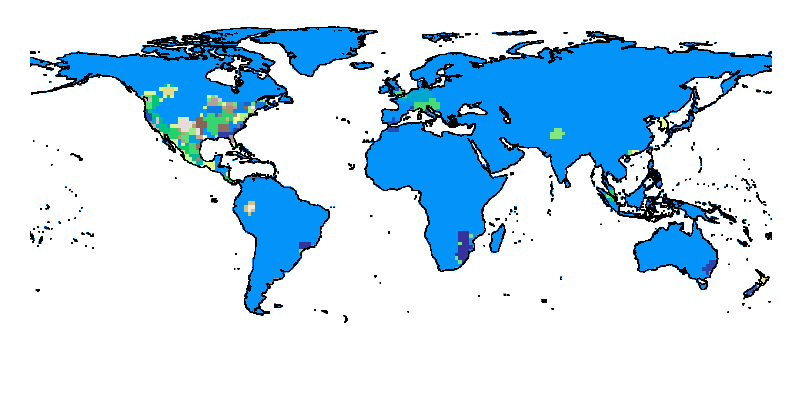}\vspaceclustering
		\caption[]{{\small 
		$\rbf$ $(\sigma = 1, \numkernel = 200)$
		}}    
		\label{fig:inat18_rbf}
	\end{subfigure}
	\hfill
	\begin{subfigure}[b]{0.33\textwidth}  
		\centering 
		\includegraphics[width=\textwidth]{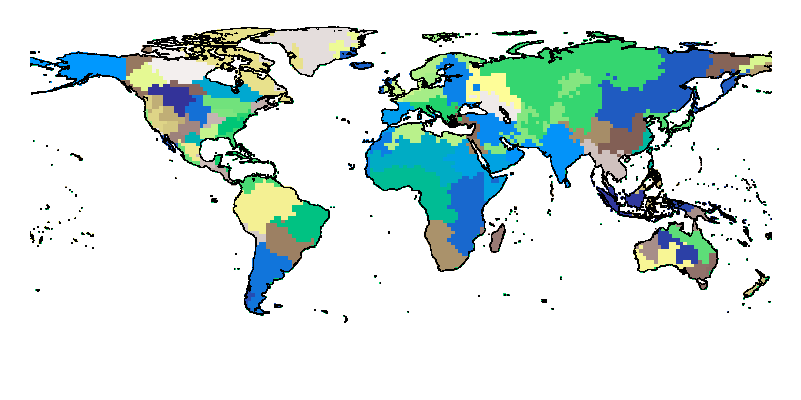}\vspaceclustering
		\caption[]{{\small 
		$\theory$ $ (\minscale = 10^{-3})$
		}}    
		\label{fig:inat18_theory_3}
	\end{subfigure}
	\hfill
	\begin{subfigure}[b]{0.33\textwidth}  
		\centering 
		\includegraphics[width=\textwidth]{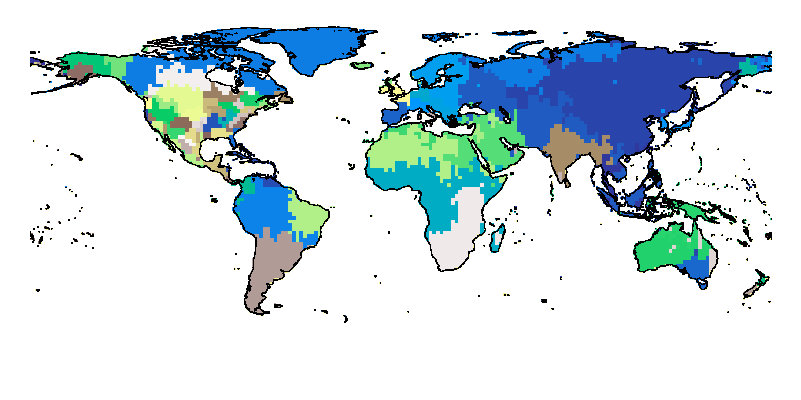}\vspaceclustering
		\caption[]{{\small 
		$\theory$ $ (\minscale = 10^{-6})$
		}}    
		\label{fig:inat18_theory}
	\end{subfigure}
	\hfill
	\begin{subfigure}[b]{0.33\textwidth}  
		\centering 
		\includegraphics[width=\textwidth]{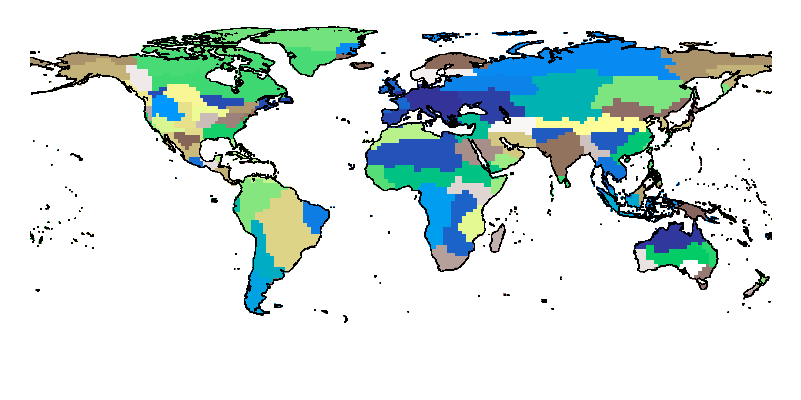}\vspaceclustering
		\caption[]{{\small 
		$\spheremixscale$ $ (\minscale = 10^{-1})$
		}}    
		\label{fig:inat18_spheremixscale_1}
	\end{subfigure}
	\hfill
	\begin{subfigure}[b]{0.33\textwidth}  
		\centering 
		\includegraphics[width=\textwidth]{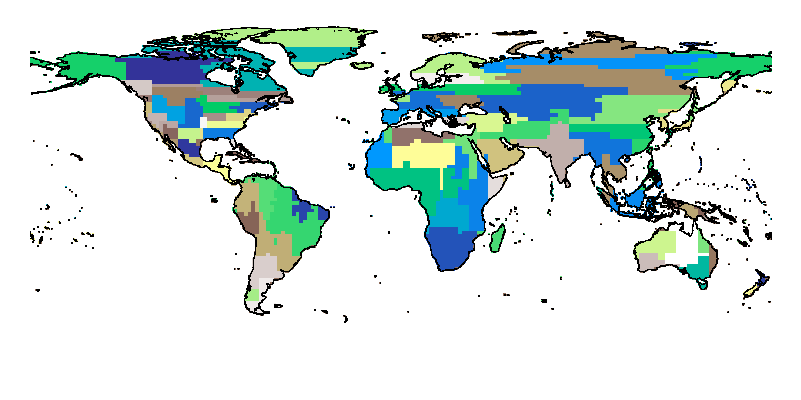}\vspaceclustering
		\caption[]{{\small 
		$\spheremixscale$ $ (\minscale = 10^{-3})$
		}}    
		\label{fig:inat18_spheremixscale_3}
	\end{subfigure}
\hfill
	\begin{subfigure}[b]{0.33\textwidth}  
		\centering 
		\includegraphics[width=\textwidth]{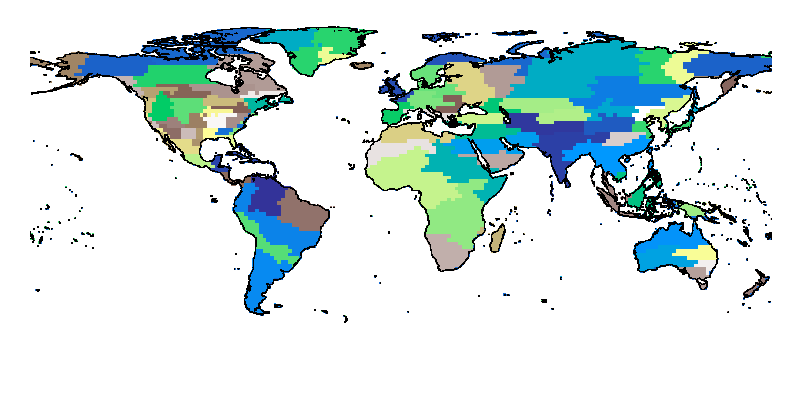}\vspaceclustering
		\caption[]{{\small 
		$\sphere$ $ (\minscale = 10^{-3})$
		}}    
		\label{fig:inat18_sphere}
	\end{subfigure}
	\hfill
	\begin{subfigure}[b]{0.33\textwidth}  
		\centering 
		\includegraphics[width=\textwidth]{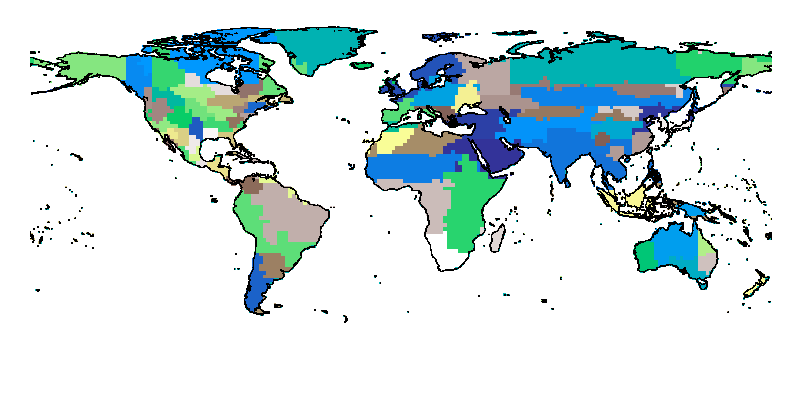}\vspaceclustering
		\caption[]{{\small 
		$\spheregrid$ $ (\minscale = 10^{-3})$
		}}    
		\label{fig:inat18_spheregrid}
	\end{subfigure}
	\hfill
	\begin{subfigure}[b]{0.33\textwidth}  
		\centering 
		\includegraphics[width=\textwidth]{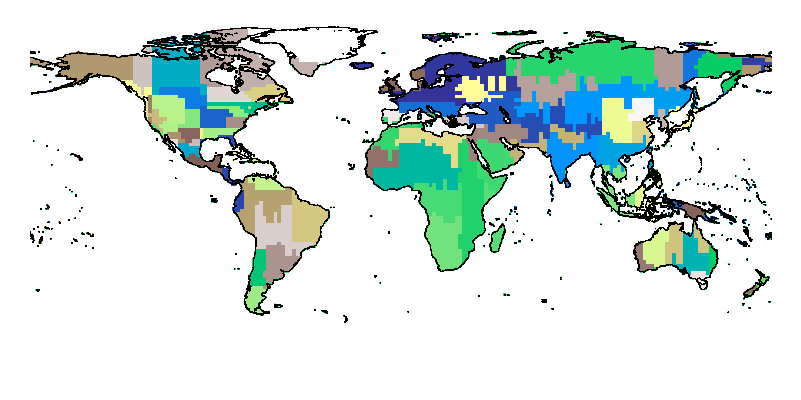}\vspaceclustering
		\caption[]{{\small 
		$\spheregridmixscale$ $ (\minscale = 10^{-3})$
		}}    
		\label{fig:inat18_spheregridmixscale}
	\end{subfigure}
	\hfill
	\begin{subfigure}[b]{0.33\textwidth}  
		\centering 
		\includegraphics[width=\textwidth]{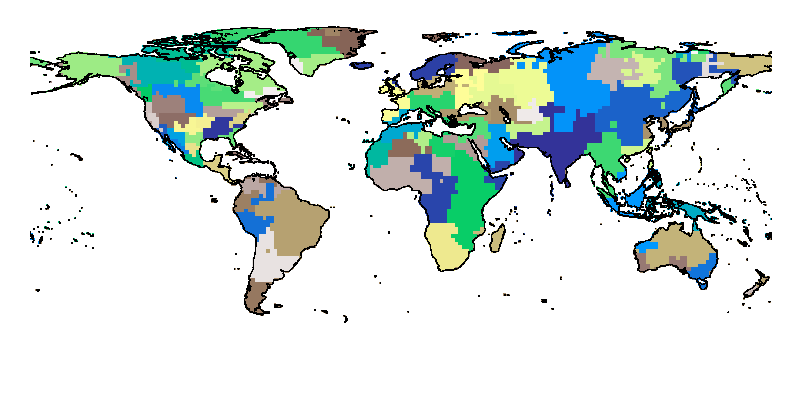}\vspaceclustering
		\caption[]{{\small 
		$\dft$ $ (\minscale = 10^{-3})$
		}}    
		\label{fig:inat18_dft}
	\end{subfigure}

	\caption[]{Embedding clusterings of iNat2018 models. 
(a) $\aodha*$ with 4 hidden ReLU layers of 256 neurons; 
	(d) $\rbf$ with the best kernel size $\kernelsize=1$ and number of anchor points $\numkernel=200$; 
(b)(c)(e)(f) are $\spacevec$ models \cite{mai2020multiscale} with different min scale $\minscale = \{10^{-6}, 10^{-3}\}$.\textsuperscript{a}
(g)-(l) are $\modelname$ models with different min scale.\textsuperscript{b}
} 
	\scriptsize
	\textsuperscript{a} {They share the same best hyperparameters:  $\freq = 64$,  $\maxscale = 1$, and 1 hidden ReLU layers of 512 neurons.} \\
	\textsuperscript{b}{They share the same best hyperparameters: $\freq = 32$, $\maxscale = 1$, and 1 hidden ReLU layers of 1024 neurons.}
	\label{fig:inat18}
\end{figure*}

We use the location encoder trained on iNat2017 or iNat2018 dataset to produce a location embedding for the center of each small latitude-longitude cell. Then we do agglomerative clustering\footnote{\url{https://scikit-learn.org/stable/modules/generated/sklearn.cluster.AgglomerativeClustering.html}} on all these embeddings to produce a clustering map. Figure \ref{fig:inat17} and \ref{fig:inat18} show the clustering results for different models with different hyperparameters on iNat2017 and iNat2018 dataset.

\end{document}